\documentclass[journal]{IEEEtran}

\usepackage{amsmath}
\usepackage{amssymb}
\usepackage[utf8]{inputenc}
\usepackage[T1]{fontenc}
\usepackage{multirow}
\usepackage{booktabs,caption}
\usepackage{adjustbox}
\usepackage{array}
\newcolumntype{L}[1]{>{\raggedright\let\newline\\\arraybackslash\hspace{0pt}}m{#1}}
\newcolumntype{C}[1]{>{\centering\let\newline\\\arraybackslash\hspace{0pt}}m{#1}}
\newcolumntype{R}[1]{>{\raggedleft\let\newline\\\arraybackslash\hspace{0pt}}m{#1}}
\usepackage{subcaption}
\usepackage{float}
\usepackage{stfloats}
\usepackage{placeins}
\usepackage[flushleft]{threeparttable}
\usepackage{gensymb}
\usepackage{siunitx}
\usepackage{tabularx,ragged2e}
\usepackage{diagbox}
\newcolumntype{C}{>{\Centering\arraybackslash}X}
\usepackage{tikz}
\usepackage{xcolor}
\usetikzlibrary{shapes, arrows.meta, positioning, calc}
\usepackage{cite}
\usepackage[hidelinks]{hyperref}

\emergencystretch=\maxdimen
\begin{document}

\title{Reliable Tree Recruitment from Scarce Annotations for Forest Inventory Expansion}

\author{Luiz~Fernando~Souza~Luz, Pedro~J.~Soto~Vega, Matheus~Pinheiro~Ferreira, Laura~Elena~Cué~la~Rosa, Hudson~Franklin~Pessoa~Veras, Ernandes~Macedo~da~Cunha~Neto, Evandro~Orfanó~Figueiredo, Ana~Paula~Dalla~Corte, Carlos~Roberto~Sanquetta, José~Matheus~Segre~Moneva~Viveiros, Matheus~Santos~Fuza, Heitor~Eduardo~Ferreira~Campos~Morato~Filpi, Cristiano~Vinícius~Vidal, Vinícius~Moura~Costa, Débora~Nascimento~Rosário, Danilo~Roberti~Alves~de~Almeida, Paulo~Guilherme~Molin, and~Dário~Augusto~Borges~Oliveira
\thanks{Corresponding author: D. A. B. Oliveira (e-mail: dario.oliveira@fgv.br).}
\thanks{L. F. Souza Luz, P. J. Soto Vega, and D. A. B. Oliveira are with the School of Applied Mathematics, Getulio Vargas Foundation, Rio de Janeiro, RJ, Brazil.}
\thanks{M. P. Ferreira and V. M. Costa are with the Department of Forest Sciences, “Luiz de Queiroz” College of Agriculture, University of São Paulo (USP/ESALQ), Piracicaba, SP, Brazil.}
\thanks{L. E. Cué la Rosa is with the Laboratory of Geo-Information Science and Remote Sensing, Wageningen University and Research, Wageningen, The Netherlands.}
\thanks{H. F. P. Veras is with the Instituto Federal do Acre, Rio Branco, AC, Brazil.}
\thanks{E. M. da Cunha Neto, A. P. Dalla Corte, and C. R. Sanquetta are with the Department of Forest Science, Federal University of Paraná (UFPR), Curitiba, PR, Brazil.}
\thanks{E. O. Figueiredo is with Embrapa Acre, Rio Branco, AC, Brazil.}
\thanks{J. M. S. M. Viveiros is with the Department of Forest Sciences, University of São Paulo (USP/ESALQ), Piracicaba, SP, Brazil, and also with Bioflore, Belo Horizonte, MG, Brazil.}
\thanks{M. S. Fuza, H. E. F. C. M. Filpi, C. V. Vidal, and D. R. A. de Almeida are with Bioflore, Belo Horizonte, MG, Brazil.}
\thanks{D. N. Rosário is with the Department of Agricultural Engineer, Federal University of Viçosa, Viçosa, Brazil.}
\thanks{P. G. Molin is with the Center for Natural Sciences, Universidade Federal de São Carlos, Buri, Brazil.}}

\markboth{Preprint}%
{Souza Luz \MakeLowercase{\textit{et al.}}: Reliable Tree Recruitment from Scarce Annotations for Forest Inventory Expansion}

\maketitle

\begin{abstract}
Species-level inventories of individual tree crowns (ITCs) from UAV imagery are limited by annotation cost: in tropical forests, many species are represented by only a handful of field-validated crowns. We propose a reliable tree recruitment framework that expands a small annotated inventory by iteratively self-training a multi-task DeepLabV3 network, combining semantic segmentation with a crown-distance regression head, on RGB UAV orthomosaics. At each iteration, candidate crowns enter the training inventory only if they pass a confidence filter on class probability and distance map and a morphology filter on crown area and solidity, with at most five new crowns per species, keeping recruited samples consistent with the reference inventory. We evaluate the framework at two Brazilian sites: an Amazonian rainforest in Acre (17 species, 355 annotated crowns, 4 cm GSD) and a forest site in the Cerrado of Maranhão (4 species, 51 annotated crowns, 1.8 cm GSD). Over 20 iterations, the number of retained crowns grew from about 500 to 960 in Acre and from about 700 to over 3100 in Maranhão, while crown-level F1 exceeded single- and multi-task DeepLabV3 baselines (53.5\% vs. 50.0\% and 45.2\%; 92.2\% vs. 79.7\% and 83.6\%). The framework offers a low-cost route to expanding species inventories with consumer-grade UAVs and limited fieldwork.
\end{abstract}


\section{Introduction}\label{Introduction}

Tropical forests are reservoirs of global biodiversity, hosting approximately two-thirds of all terrestrial species \cite{gardner2009prospects}. Their vastness and inaccessibility have historically posed challenges for biodiversity monitoring, making it difficult to understand how global changes affect their capacity to harbor species. However, advances in high-resolution remote sensing are ushering in a new era of tropical forest monitoring \cite{fassnacht2016review}. UAVs, capable of capturing imagery with centimeter-level spatial resolution, enable species detection at the individual tree crown (ITC) level \cite{ferreira2020individual, veras2022fusing}. This detail transforms biodiversity assessment, allowing researchers to map and monitor tree species distributions, even in tropical forests' dense and complex canopies.

Current approaches for tree species mapping are based on deep learning methods, with convolutional neural networks (CNNs) standing out in this field \cite{kattenborn2019convolutional}. CNNs are an artificial neural network designed to process and analyze image data by automatically learning spatial hierarchies of features through convolutional layers. These networks can capture species-specific crown structural properties in high-resolution images, enabling accurate species classification. \cite{ferreira2020individual} is one of the pioneering studies to identify tree species in Amazonian forests using UAV imagery and CNNs. The authors employed images with a ground sample distance (GSD) of 4 cm and reported an average accuracy of 87.8\% in distinguishing four palm tree species. \textit{Euterpe precatoria}, known as Açaí, a palm of economic importance, was mapped with over 98\% accuracy. \cite{veras2022fusing} used a CNN for semantic segmentation and UAV images acquired across different seasons to classify Amazonian tree species. They found that combining multi-season images improved species classification, reaching an average accuracy of 90.5 \% in classifying eight tree species. Nonetheless, these studies were conducted in areas with relatively low tree species diversity and relied on datasets with more than 25 ITCs per species, which may have contributed to the high classification accuracies reported. 

However, obtaining sufficient reference ITCs for training remains a challenge in tropical forests due to the difficulties of field data collection and the high diversity of species. While successful studies in controlled environments or with more common species can benefit from hundreds or thousands of labeled samples per class, the reality for many species in tropical forests is the availability of only a handful of labeled ITCs. This scarcity is especially pronounced for rare species, which are typically represented by only a few training samples \cite{mellor2015exploring, fassnacht2016review}. Increasing the set of samples from rare species may address this issue, but this alternative requires time and proves to be costly, as shown by \cite{graves2016tree1}. This data disparity limits the application of deep learning models that rely on large volumes of annotated data to achieve high performance and generalization \cite{zhu2017deep, ma2019deep}.

Some innovative approaches have been developed to address the scarcity of labeled data for tree species classification in tropical forests. For instance, \cite{cuerosa2021multi} proposed a multi-task fully convolutional network (MTFCN) to classify tree species in dense tropical forests using sparse polygon-level annotations from hyperspectral UAV data. Their method incorporates a partial loss function to train the network with non-dense annotations. It introduces a complementary task for distance regression, which enforces tree crown boundary constraints and enhances classification performance. The MTFCN achieved an average user’s accuracy of 88.63\% and producer’s accuracy of 88.59\%, improving results for underrepresented species with fewer than three ITCs in the dataset.

Another promising approach for improving tree species mapping is active learning \cite{settles2012active, tuia2011survey}. In an active learning pipeline, the model is initially trained on a small annotated dataset and then used to make predictions on a larger pool of unlabeled data. It then makes predictions on a larger pool of unlabeled data, identifying areas or samples where additional annotations enhance model performance. 
Typically, the selected samples are those where the model exhibits the highest uncertainty or faces the most significant difficulty in making accurate predictions \cite{ren2021survey}. The selected samples are then submitted to an oracle, an agent responsible for providing new annotations to enrich the annotated dataset and improve the initial model. Usually, the role of the oracle is performed by a human annotator with knowledge of the domain of the data. Then, the new labels are added to the training set to start a new cycle of the method. This iterative process gradually improves the model's performance until a stopping criterion is reached.

In \cite{kellenberger2019half}, the authors present a deep active learning strategy for object detection in UAV-obtained images, aiming to map wildlife in savanna areas. In contrast to traditional active learning techniques, in \cite{kellenberger2019half}, the oracle is requested to label the set where the model is most confident about its predictions. This strategy aims to reduce the number of false positives common in scenarios with rare or unbalanced classes. It aims to minimize the annotator's efforts, who can focus on the most relevant cases. The authors show that the proposed technique achieves reasonable precision indicators, even with a lower annotation cost than conventional active learning methods. In \cite{tong2024active}, the authors propose an active learning-based spectral-spatial classification (ALSSC) model for tree species classification in hyperspectral images. The ALSSC employs multiscale superpixels to extract spectral-spatial features and applies guided filtering to refine class probability maps while preserving edge information. Active learning is then integrated to iteratively select the most informative samples based on uncertainty, enriching the training set after each classification round. The proposed ALSSC achieved an overall accuracy of 94.86\%. However, the ALSSC relies on hyperspectral imagery and performs classification at the pixel/superpixel level, rather than detecting and classifying individual tree crowns from the RGB UAV imagery used in our study.

The examples above demonstrate the potential of active learning methods to address challenges in data-scarce scenarios and achieve high classification accuracy. However, these studies were conducted in environments with limited species diversity, such as savanna areas or controlled datasets. Despite their potential, active learning approaches have not been extensively tested in tropical forest environments, where the high species diversity and structural complexity, including dense and overlapping canopies, pose unique challenges for ITC level segmentation and classification. Thus, exploring the applicability of these methods in such biodiverse ecosystems remains an important area of research. Moreover, these approaches still rely on a human oracle to label the selected samples. When the scarce resource is botanical expertise in the field, a natural question is whether the model itself can act as the oracle, that is, whether reliable pseudo-labels can be selected automatically in a self-training fashion, without new manual annotation, to identify tree species at the ITC level.

In this work, we address the problem of expanding forest species inventories from a limited number of annotated individual tree crowns (ITCs). We propose a reliable tree recruitment framework that progressively enlarges forest inventories through the automatic incorporation of high-confidence tree crowns validated by crown morphology. The proposed framework is implemented through an iterative self-training (pseudo-labeling) strategy combining semantic segmentation, confidence estimation and morphology-aware pseudo-label validation. By progressively expanding the inventory, the proposed approach improves species detection and classification while substantially reducing manual annotation requirements.

The main contributions of this work:

\begin{enumerate}
    \item We propose a framework for reliable tree recruitment that progressively expands forest species inventories from a small set of annotated individual tree crowns while preserving inventory quality.

    \item We introduce a morphology-aware recruitment strategy that combines confidence estimation and crown morphology validation to automatically identify reliable candidate trees.

    \item We implement the proposed recruitment strategy through a multi-task semantic segmentation framework and demonstrate consistent improvements across two tropical forest datasets.
\end{enumerate}

This approach supports scalable biodiversity monitoring by enabling the model to act as an automated oracle for label generation, reducing the reliance on manual annotation.

The remainder of this paper is organized as follows. Section \ref{Materials} presents a detailed description of the proposed tree recruitment  framework. Section \ref{Experimental Design} describes the training and validation procedures as well as the datasets used in this study. Section \ref{Results} reports the performance of the proposed method under different scenarios. Finally, Section \ref{Conclusion} summarizes the main findings and outlines directions for future work.

\section{Deep Automatic Tree Recruitment  for Forest Species Mapping using Multitask Semantic Segmentation}
\label{Materials}

\begin{figure*}[t]
    \centering
    \includegraphics[width=0.85\textwidth]{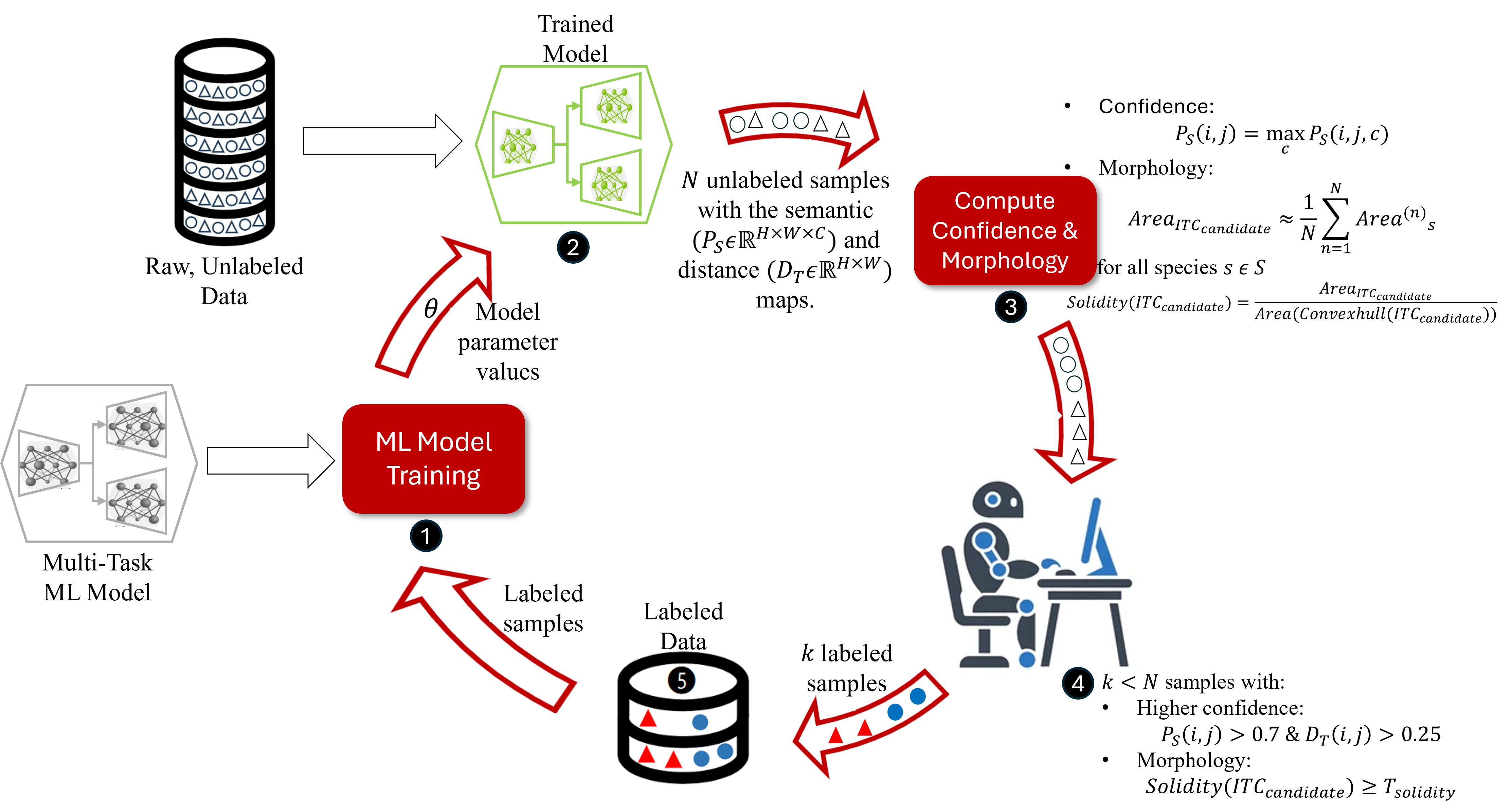}
    \caption{Overview of the proposed framework. (1) Training the ML model with the labeled samples. (2) Predicted semantic segmentation and the distance map by the multi-task model. (3) Prediction refined with a confidence threshold above 70\% and distance map value above 0.25. (4) Final Individual Tree Crowns (ITCs) selected and prepared for the tree recruitment  scheme's sample inventory, incorporating morphological validation.}
    \label{fig:model_overview}
\end{figure*}

 This section presents the proposed reliable tree recruitment framework for progressive forest inventory expansion under scarce annotations. As illustrated in Figure \ref{fig:model_overview}, the framework follows an iterative process in which newly recruited tree crowns are automatically incorporated into the inventory after confidence and morphology validation. The recruitment strategy is implemented through a self-training workflow combining multi-task semantic segmentation and iterative pseudo-label refinement.

 The workflow begins with an initial set of labeled data, which trains a multi-task deep learning model. This model then processes unlabeled data, its predictions to identify high-confidence samples. An aspect of this process involves the selection of pseudo-labels, refined not only by model confidence and distance transform outputs but also validated through crown morphology heuristics. These selected samples are then integrated into the training dataset, improving the model's ability to identify and classify ITCs with reduced reliance on extensive manual annotation. This integrated approach aims to ensure generalization and high classification accuracy across diverse species, including those that are underrepresented.

\subsection{Overall Multi-Task Architecture}

\begin{figure*}[t]
    \centering
    \includegraphics[width=0.95\textwidth]{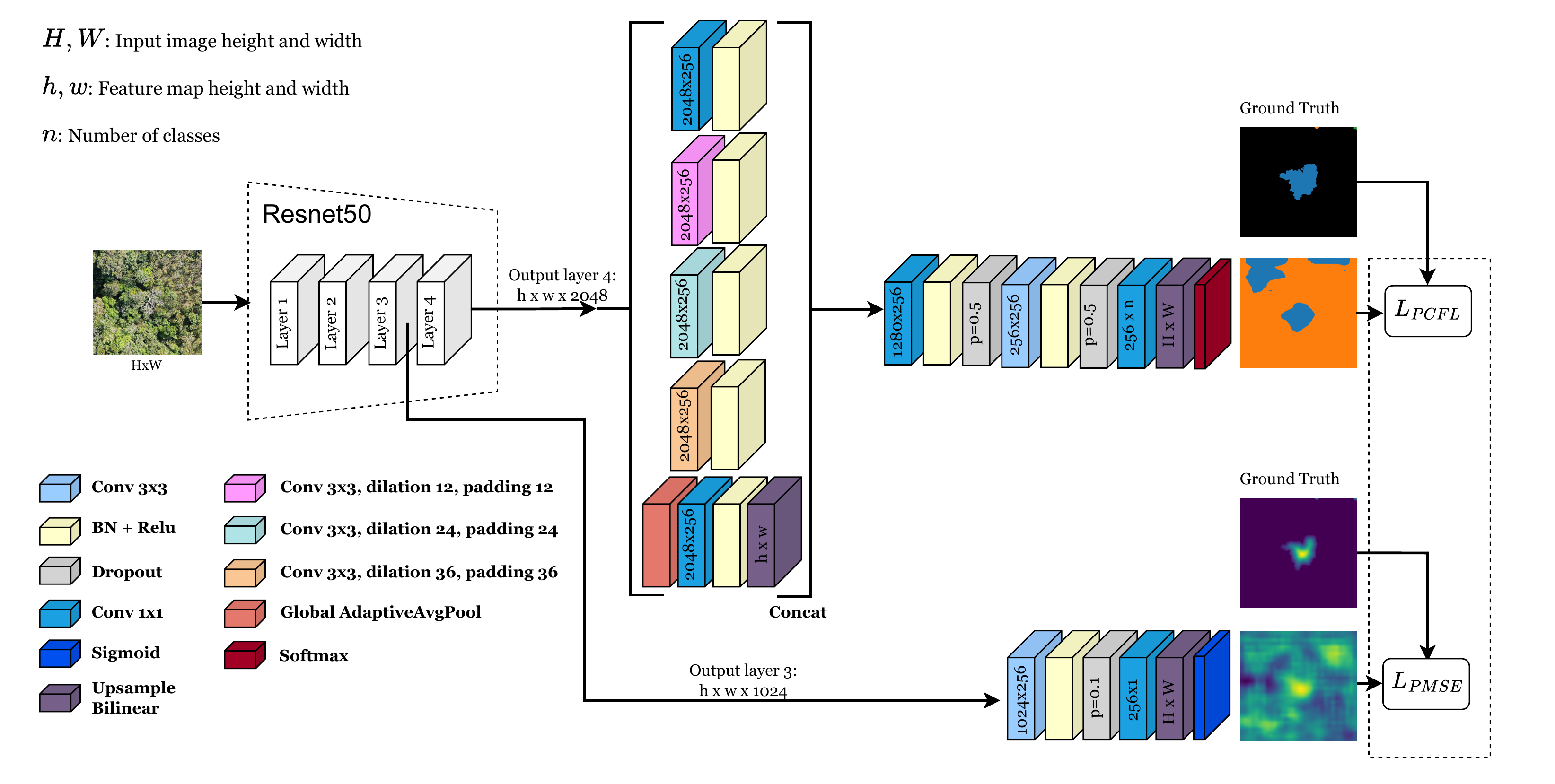}
    \caption{Overview of the proposed multitask DeepLabV3 architecture. It features a shared encoder, built upon a ResNet-50 backbone, that branches into two specialized decoders: one for semantic segmentation (supervised by $L_{PCFL}$) and another for distance transform map prediction (supervised by $L_{PMSE}$).}
    \label{fig:multitask_architecture}
\end{figure*}

As depicted in Figure \ref{fig:model_overview} (step 1), the tree recruitment  cycle begins with the training of a ML model which consist in a deep learning (DL)-based multi-task framework. Figure \ref{fig:multitask_architecture} provides a conceptual overview of this multi-task architecture. Following \cite{cuerosa2021multi}, the multi-task network aims at producing two different outputs by processing an input UAV image. The first prediction is a semantically labeled image (representing the main classification task) and the second is a distance transform map (serving as an auxiliary regression task). The architecture comprises a shared encoder network and two specialized decoder networks, each tailored to its respective task. This design capitalizes on the ability of the shared encoder to extract hierarchical features common to both tasks, while the separate decoders allow for task-specific optimization. The synergy between these tasks enables the model to learn more robust features and improve the precision of individual tree crown boundaries.

\subsubsection{Semantic segmentation as main task}
\label{Instance Segmentation as Main Task}

The semantic segmentation branch relied on the well known DeepLabV3 \cite{chen2017rethinking} architecture with a backbone based on ResNet \cite{he2016deep} (see Figure \ref{fig:multitask_architecture}). To address the challenges related to sample imbalance and data sparsity, the adopted strategy consisted of using the Partial Categorical Focal Loss (\emph{$L_{PCFL}$}), as it has shown satisfactory results in \cite{cuerosa2021multi}. The \emph{$L_{PCFL}$} incorporates two fundamental loss concepts: Partial Loss \cite{tang2018normalized, wu2018scribble} and Focal Loss \cite{lin2017focal}. This function can be defined by the following:

\begin{equation}\label{equ:LPCFL}
L_{PCFL} = -\frac{1}{|\Omega_{L}|} \sum_{i\in\Omega}\omega_{i}\sum_{j \in C}y_{ij}(1 - \hat{y}_{ij})^{\gamma} \log{(\hat{y}_{ij})}
\end{equation}

Here, $\Omega_{L}$ represents the set of pixels with annotation, $|\Omega_{L}|$ the total number of pixels in this set, $C$ denotes the set of possible classes, $y_{i,j}$ is a binary variable indicating the membership of pixel $i$ to class $j$, $\hat{y}_{i,j}$ is the estimated probability by the model that pixel $i$ belongs to class $j$, and $\omega_{i}$ is an indicator variable of $i\in{\Omega_{L}}$. Finally, the parameter $\gamma \geq 0$ represents the loss's focalization degree.

In the development of this loss function, the concept of partial loss introduced by \cite{wu2018scribble} and \cite{tang2018normalized} was used. In this construction, the network weights are updated exclusively based on the set of pixels $i$ belonging to the annotated set $\Omega_{L}$. It was demonstrated in \cite{wu2018scribble} that this strategy improved results in sparse datasets.

The concept of focal loss, as presented by \cite{lin2017focal}, is also incorporated. In this approach, the loss function focuses on hard-to-classify cases and reduces the importance of correctly classifying cases with high model confidence. The parameter $\gamma$ defines the degree of focus the loss should apply to hard-to-classify cases. Note that when $\gamma = 0$, this function corresponds to the traditional categorical cross-entropy. The incorporation of focal loss is justified by its effectiveness in addressing data imbalance problems compared to traditional categorical cross-entropy.

\subsubsection{Distance transform as auxiliary task}

In this work, a method similar to that proposed by \cite{cuerosa2021multi} was utilized to enhance the generalizability of the network. This method integrates an auxiliary task into the model, serving as a regularizer. Here, the network is trained to learn a regression function that maps the input image to a distance map transform. This approach has been proven to be effective in applications such as mapping tree species \cite{martins2021deep, lassalle2023advances} and automatically delineating tree crowns \cite{lassalle2022deep} in tropical environments.

To construct the training set for the distance map transform, the labels of each ITC underwent several processing steps. Initially, binary masks that corresponded to each ITC were generated. Subsequently, a distance map, denoted as \emph{d}, was calculated for each component of the binary image using the euclidean distance transform (EDT). This transformation produces a distance map where the value assigned to each internal pixel of the instances indicates the Euclidean distance to the nearest boundary.

\begin{figure}[h]
	\centering
	\includegraphics[width=\linewidth]{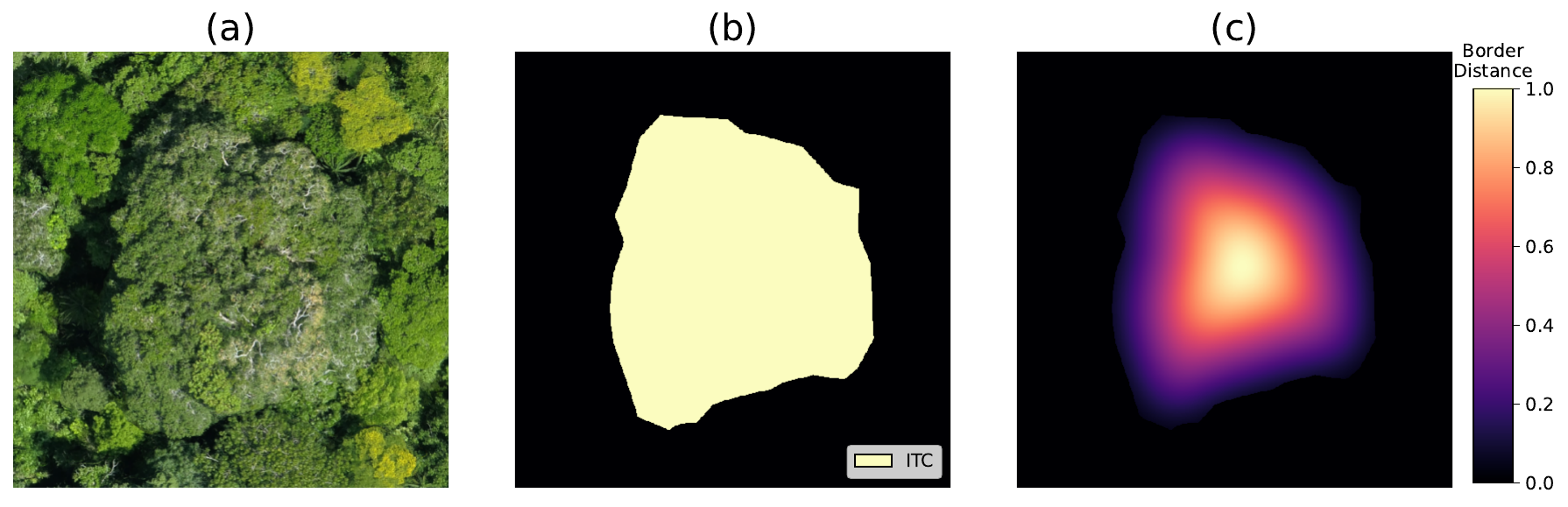}
	\caption{Example of the distance transform generation.
	(a) RGB image.
	(b) Binary mask of the ITC.
	(c) Euclidean distance transform with Gaussian filter}
	\label{fig:figure2}
\end{figure}

A Gaussian filter is applied to smooth the distance map transformation, primarily to mitigate noise and irregularities in the data. The selected variance, $( \sigma = 30)$, strikes a reasonable balance between retaining essential features in the distance map and eliminating unwanted noise. This specific value for $\sigma$ was determined empirically, based on visual inspection and preliminary tests that showed optimal smoothing for capturing crown morphology variations without over-blurring critical boundary information. Due to the variability in canopy sizes—even among trees of the same species—normalization of the distance transform was necessary for each ITC. This normalization rescales the maximum values for each ITC to 1, so that they refer to the center of the crown, as illustrated in Figure \ref{fig:figure2}c.

According to the work of \cite{cuerosa2021multi}, incorporating the distance map transform into the classification process causes the network to predict lower values near the edges of tree crowns and higher values next to the centre. This constraint helps to more accurately delineate ITCs, including those classified under the same semantic group. This method highlights the boundaries between adjacent tree crowns, improving the model's ability to distinguish between closely situated trees.

As in the semantic segmentation block, a partial loss function was employed to train the distance map \emph{decoder}. The problem was modeled as a regression problem, using the $L_2$ loss function (mean squared error - MSE). The MSE represents the sum of squared distances between the reference map and the estimated map, and the modified version of this function was done as follows:

\begin{equation}
    L_{PMSE} = \dfrac{1}{|\Omega_{L}|}\sum_{i \in \Omega} \omega_{i} (d_{i} - \hat{d}_i)^{2}
\end{equation}

Where $\omega_{i}$ is an indicator variable of pixel $i \in \Omega_{L}$, $d_{i}$ is the distance from pixel $i$ to the polygon edge, and $\hat{d}_{i}$ is the estimate generated by the network.

\subsubsection{Multi-task loss function}

The final \emph{Loss} $L$ of the model can be defined by the formula:

\begin{equation}
    L = L_{PCFL} + \lambda L_{PMSE}
\end{equation}

As mentioned above, the loss function $L_{PCFL}$ influences the main classification task of the network. In contrast, $L_{PMSE}$ plays the role of a network regularizer at the same time that it helps to separate ITCs that pertain to the same category. 
The coefficient $\lambda$ indicates the intensity of the regularization applied. For the experiments conducted here, $\lambda = 1$ was established. This value was chosen after a series of preliminary experiments where different $\lambda$ values were tested, and $\lambda = 1$ consistently provided the best balance between the main classification task and the auxiliary regularization task, preventing either task from dominating the learning process.

\subsection{Automatic Tree Recruitment  Methodology}
\label{sec:active_learning_methodology}

This section details the proposed tree recruitment  methodology, an iterative, model-guided approach designed to efficiently generate high-quality pseudo-labels to address the high cost of manual annotation. The complete tree recruitment  cycle is illustrated in Figure \ref{fig:model_overview}.

\subsubsection{Dense Prediction over the Study Area}
After initial training, the trained multi-task model is applied to the entire study area (top cylinder in Figure \ref{fig:model_overview}) to generate dense pixel-wise predictions (Step 2). This process yields a likelihood map $P_S \in \mathbf{R}^{W \times H \times C}$ (indicating class membership probability for each pixel) and a distance map $D_S \in \mathbf{R}^{W \times H}$ (encoding the distance of each pixel to the nearest ITC boundary) over the full orthoimage. Individual ITC candidate components are then extracted from these dense predictions for subsequent filtering.

\subsubsection{Confidence and Morphology Criteria}
Following the prediction stage, two key criteria are computed to select the most informative pseudo-labeled samples for further training (Step 3 in Figure \ref{fig:model_overview}): confidence and morphology.

\textit{Confidence-based Criteria:}
This criterion exploits the probability maps $P_S$ and the distance map $D_S$. Pixels from the segmented images are selected if their class membership probability indicates high certainty, specifically when classified with at least 70\% confidence. This threshold balances the number of pixels included with the reliability of their classification. The distance map $D_S$ further refines ITC delineation through a 0.25 threshold on its normalized values, set empirically after testing values from 0.10 to 0.30 on the validation set. Below this threshold, residual boundary noise degraded the delineation; above it, the crowns were over-eroded.

\textit{Morphology-based Criteria:}
Only instances that morphologically resemble an ITC are selected. This involves two sub-criteria:
\begin{itemize}
    \item \textbf{Crown Area:} Instances with crown areas closest to their respective species' median area are chosen. Experimental results demonstrate that the median identifies representative ITCs.
    \item \textbf{Solidity:} A filter is applied using the \emph{solidity} metric, defined as the ratio of a geometric object's area to its convex hull's area. This metric quantifies how closely an instance approximates a convex shape. Components deviating from a convex form are discarded, aligning with the observed convex and rounded shapes of ITCs in the initial training set.
\end{itemize}
These selection steps ensure that only reliable and morphologically sound samples are considered for subsequent training.

\subsubsection{Storage and Selection from Sample Inventory}
The validated pseudo-labeled ITCs, selected based on confidence and morphology criteria, are then considered and stored in a sample inventory (Step 4 in Figure \ref{fig:model_overview}). This inventory accumulates reliable components across iterations. If a sample already exists at a specific location, the older instance is replaced by the newly obtained, potentially refined shape from the most recent iteration. This replacement applies only to pseudo-labels within the inventory. The manually annotated ITCs from the initial training set remain unchanged throughout all iterations and are never overwritten by model predictions.

\subsubsection{Training Set Update and Iteration}
From the sample inventory, a maximum of five new ITCs per species (not previously part of the training data) are selected to update the current training set (Step 5 in Figure \ref{fig:model_overview}). This species-balanced selection aims to maintain data stability and prevent overrepresentation of certain species. The updated training set, which now includes both the original labeled data and the newly selected high-confidence pseudo-labels, is then used for the next iteration of model training, restarting the cycle at Step 1. This iterative expansion of the training set with diverse and informative samples improves the model's ability to identify and classify ITCs, leading to increased detection of ITCs per species while maintaining high classification accuracy. 

The detection step is the same at every iteration, including the final one: the confidence thresholds ($P_S \geq 70\%$ and normalized $D_S \geq 0.25$) and the morphology criteria (crown area close to the species' median and the solidity filter described in Section~\ref{sec:active_learning_methodology}) are applied unchanged. The final inventory reported in the experimental analysis corresponds to the state of the sample inventory at the last iteration (e.g., the 13th and 20th iterations referenced in Figures \ref{fig:f1score_nr} and \ref{fig:f1score_allspecies}), obtained by applying the model trained in that iteration to the entire study area and filtering its dense predictions through the same confidence and morphology pipeline.

\FloatBarrier
\section{Experimental analysis}
\label{Experimental Design}

This section details the experimental design and setup developed to evaluate the performance of the proposed deep tree recruitment  framework. It encompasses a description of the study areas, the methodology for UAV image acquisition, the characteristics of the individual tree crown (ITC) datasets, and the specific configuration of the multi-task CNN architecture. Furthermore, the section outlines the training protocols, including data augmentation and optimization strategies, and the metrics used for model evaluation.

\subsection{Study Areas}

\begin{figure*}[t]
    \centering
    \includegraphics[width=0.85\textwidth]{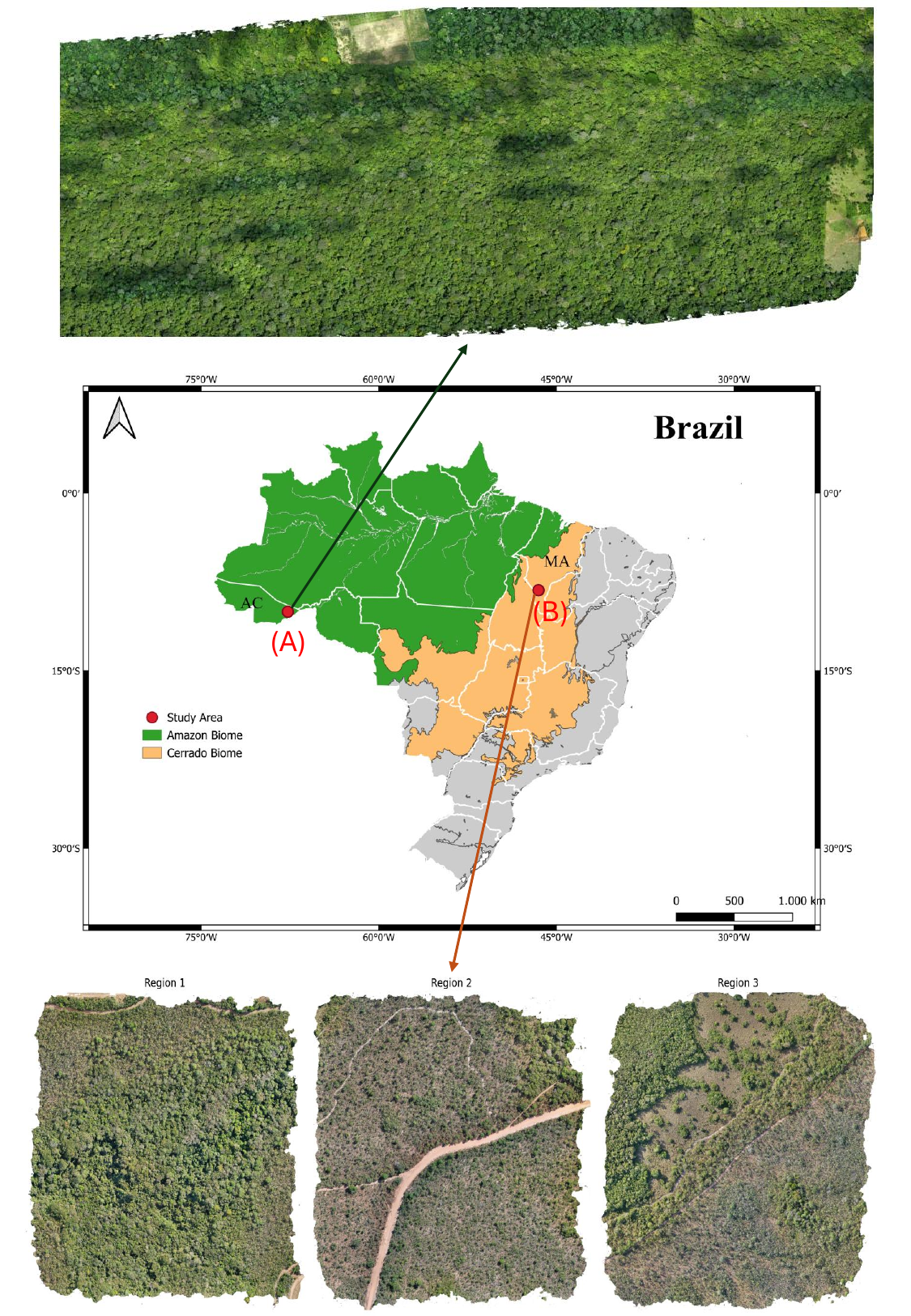}
    \caption{Study areas location. (A) Embrapa experimental forest in Rio Branco, Acre (AC) state, Brazil. (B) Baixao Preto Farm in Riachao, Maranhao (MA) state, Brazil, showing 3 of the 15 surveyed regions.}
    \label{fig:study_areas}
\end{figure*}

Two regions were selected to evaluate the performance of the proposed method, as shown in Figure \ref{fig:study_areas}. The first, region (A), is an experimental forest of the Brazilian Agricultural Research Company (Embrapa), located in the municipality of Rio Branco, Acre state, Brazil ($10^\circ 01' 22'' \, \text{S}, 67^\circ 40' 3'' \, \text{W}$). The site comprises 1,600 hectares of highly diverse Amazonian forest, approximately 200 meters above sea level. The specific area analyzed in this study covered 260 hectares. This specific 260-hectare area was selected due to its representative diversity species and its accessibility for establishing ground control points and conducting detailed field inventories, which were crucial for ground-truthing the UAV imagery and individual tree crown identification. The climate is characterized by an average annual rainfall of 1,950 mm and a mean annual temperature of $24.8 \pm 0.8^\circ \text{C}$. The vegetation is classified as open rain-forest. The forest structure includes trees exceeding 30 meters in height, abundant bamboo, diverse palm species, and large emergent trees.

Over 239 tree species have been documented in this forest. The canopy is multi-layered, with some individual trees reaching heights of up to 40 meters. The diameter at breast height (DBH) of trees included in the study ranged from 50 cm to 190 cm, with a distribution typical of a mature, unmanaged tropical forest. The challenging nature of this environment, with its dense canopy and spatial arrangements, offers a test for the capabilities of the proposed approach.

The second study area, highlighted by (B) in Figure \ref{fig:study_areas}, is a Baixao Preto Farm. Located in 42.29 $W$ and 8.14$S$, municipality of Riachao, southern of Maranhao State, Brazil. The farm spans 6273.49 hectares within the Cerrado biome, primarily used for agriculture, accounting for 61.88\% of land use land cover. Additionally, it includes Alluvial Semideciduous Seasonal Forest (13.13\%), Wooded Savanna (15.35\%), Park Savanna (8.64\%), and non-vegetated areas (1.04\%). The local climate is a tropical savanna with dry winters according to the classification system of Koppen.

\subsection{Datasets}

In region (A), aerial photographs were collected in November 2017 (rainy season) using a DJI Phantom 4 Pro unoccupied aerial vehicle (UAV) equipped with an RGB (red, green, blue) CMOS sensor. The sensor had a resolution of 20 megapixels and was fitted with a 24 mm autofocus lens and manual shutter. 
The aerial survey was conducted at a flight altitude of 120 meters above the forest canopy, with the UAV maintaining a cruising speed of 10 \, \text{m/s}. This configuration resulted in images with a ground sampling distance (GSD) of 4 \, \text{cm}, providing high-resolution forest canopy data. Before the flights, three ground control points (GCPs) were established on the edges of the forest reserve to ensure geometric accuracy for the entire area. These GCPs were surveyed using dual-frequency GNSS receivers, collecting GPS and GLONASS data for 241 minutes. Post-processing of the GCP data yielded an average vertical precision of 10\, \text{cm} and a horizontal precision of 3\, \text{cm}. This configuration, combined with the flight plan, ensured optimal geometric correction and minimal distortion across the reconstructed orthomosaic. The raw UAV images were then processed using structure-from-motion photogrammetry software to generate a high-resolution orthomosaic and a Digital Surface Model (DSM). Standard photogrammetric workflows were applied, including bundle adjustment and dense cloud generation, followed by orthorectification and mosaicking to produce the final imagery used for analysis.

On the other hand, in region (B), a total of 254 hectares with native vegetation were surveyed in July 2023, during the dry season, using a DJI Mavic 2 Pro (M2P) UAV. The drone was flying at a height of 120 meters with 85 \% front and side overlap, resulting in 15 regions of 1.8 cm/pixel resolution and three channels (RGB). Although these regions are essentially dense forests, they also have parts covered by roads and plantations. These areas can be observed in Figure \ref{fig:study_areas} where 3 of the 15 regions are represented. Simultaneously with the drone imagery, field data was collected in the same period from 45 plots each 900$m^2$ in size, where 90 species belonging to 36 families were identified. Among these, 8 species were palms, and 82 had tree or shrub habits. However, only four species were selected for model training and evaluation. Additionally, 238 trees of 52 species were geo-referenced using a Global Navigation Satellite System (GNSS) L1/L2 device. All the annotations have their delimited segmentation inside the region surveyed by drones.

\begin{table*}[t]
\centering
\caption{Summary of ITCs and pixel data for different tree species in the Amazon AC dataset.}
\begin{adjustbox}{width=1\textwidth,center}
\begin{tabular}{llllllll}
\hline
\multicolumn{1}{c}{\multirow{2}{*}{\textbf{Species names}}} & \multicolumn{1}{c}{\multirow{2}{*}{\textbf{Scientific names}}} & \multicolumn{2}{c}{Total ITCs} & \multicolumn{2}{c}{Training ITCs} & \multicolumn{2}{c}{Testing ITCs} \\
\multicolumn{1}{c}{} & \multicolumn{1}{c}{} & \multicolumn{1}{c}{\textbf{Nº of crowns}} & \multicolumn{1}{c}{\textbf{Nº of pixels (M)}} & \multicolumn{1}{c}{\textbf{Nº of crowns}} & \multicolumn{1}{c}{\textbf{Nº of pixels (M)}} & \multicolumn{1}{c}{\textbf{Nº of crowns}} & \multicolumn{1}{c}{\textbf{Nº of pixels (M)}} \\ \hline
Abiorana Rosa & \textit{Micropholis venulosa} & 72 & 16.37 & 50 & 11.4 & 22 & 4.97 \\
Angico & \textit{Parkia nitida} & 6 & 1.11 & 4 & 0.898 & 2 & 0.216 \\
Angico Vermelho & \textit{Parapiptadenia rigida} & 14 & 3.32 & 9 & 1.88 & 5 & 1.44 \\
Castanheira & \textit{Bertholletia excelsa} & 83 & 15.65 & 58 & 10.78 & 25 & 4.87 \\
Cedro & \textit{Cedrela odorata} & 52 & 3.91 & 36 & 2.54 & 16 & 1.37 \\
Cerejeira & \textit{Amburana acreana} & 14 & 0.95 & 9 & 0.66 & 5 & 0.29 \\
Cumaru Ferro & \textit{Dipteryx odorata} & 20 & 4.46 & 14 & 3.20 & 6 & 1.26 \\
Garapeira & \textit{Apuleia leiocarpa} & 45 & 5.21 & 31 & 3.26 & 14 & 1.95 \\
Guaribeiro & \textit{Brosimum rubescens} & 24 & 2.01 & 16 & 1.49 & 8 & 0.53 \\
Guariuba & \textit{Clarisia racemosa} & 18 & 1.22 & 12 & 0.80 & 6 & 0.42 \\
Ipe & \textit{Handroanthus impetiginosus} & 20 & 1.06 & 14 & 0.74 & 6 & 0.32 \\
Jutai & \textit{Hymenaea oblongifolia} & 32 & 3.69 & 22 & 2.51 & 10 & 1.17 \\
Massaranduba & \textit{Manilkara huberi} & 9 & 1.13 & 6 & 0.64 & 3 & 0.49 \\
Samauma & \textit{Ceiba pentandra} & 11 & 4.42 & 7 & 2.69 & 4 & 1.74 \\
Tachi & \textit{Tachigali spp.} & 25 & 3.35 & 17 & 2.45 & 8 & 0.90 \\
Tauari & \textit{Couratari macrosperma} & 67 & 9.97 & 46 & 6.81 & 21 & 3.16 \\
Ucuuba & \textit{Virola surinamensis} & 6 & 0.41 & 4 & 0.33 & 2 & 0.08 \\ \hline
\end{tabular}
\end{adjustbox}
\label{tab:tree_species_data}
\end{table*}

\subsubsection{Individual tree crown database (Amazon AC)}

This study utilized an enhanced version of the ITC database, developed by \cite{veras2022fusing}. This version includes economically significant tree species for both timber and non-timber products. The ITCs were delineated from UAV images, and trees were identified at the species level using forest inventory data and expert botanical knowledge. Table \ref{tab:tree_species_data} summarizes the number of ITCs and pixels per species.

\begin{figure*}[t]
	\centering
	\includegraphics[width=0.95\textwidth]{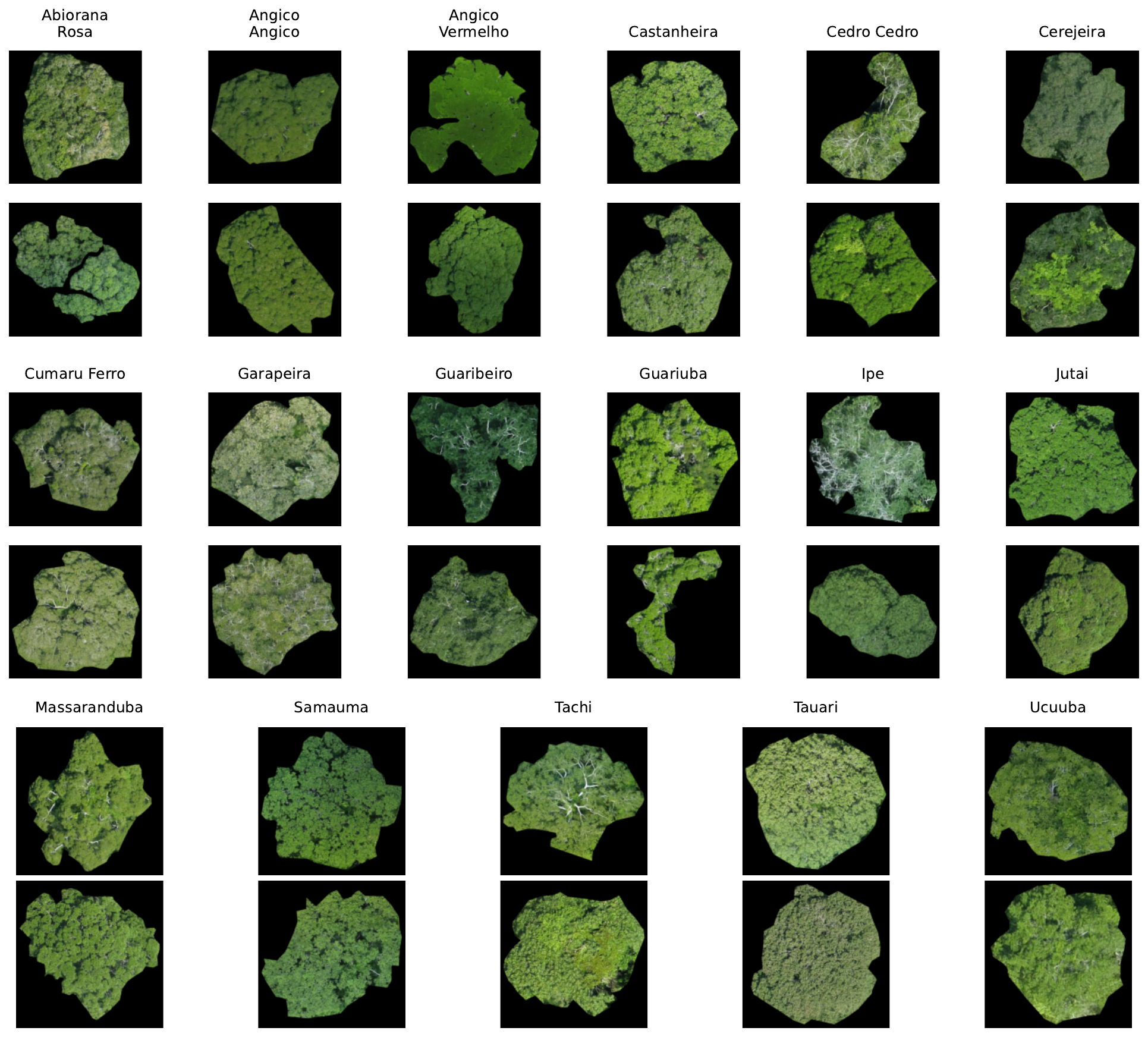}
	\caption{Examples of individual tree crowns of the 17 target species in the Amazon AC dataset (UAV RGB imagery, 4 cm GSD).}
	\label{fig:figure_seventeen_species_contrasting_crowns}
\end{figure*}

Figure \ref{fig:figure_seventeen_species_contrasting_crowns} presents an example of the UAV image showcasing the comprised tree species, with contrasting crown architectures. The UAV image, with a GSD of 4 cm, provides high-resolution detail to capture species-specific differences in crown structural patterns.

\subsubsection{Individual tree crown database (Bioflore MA)}

\begin{figure*}[t]
	\centering
	\includegraphics[width=0.7\textwidth]{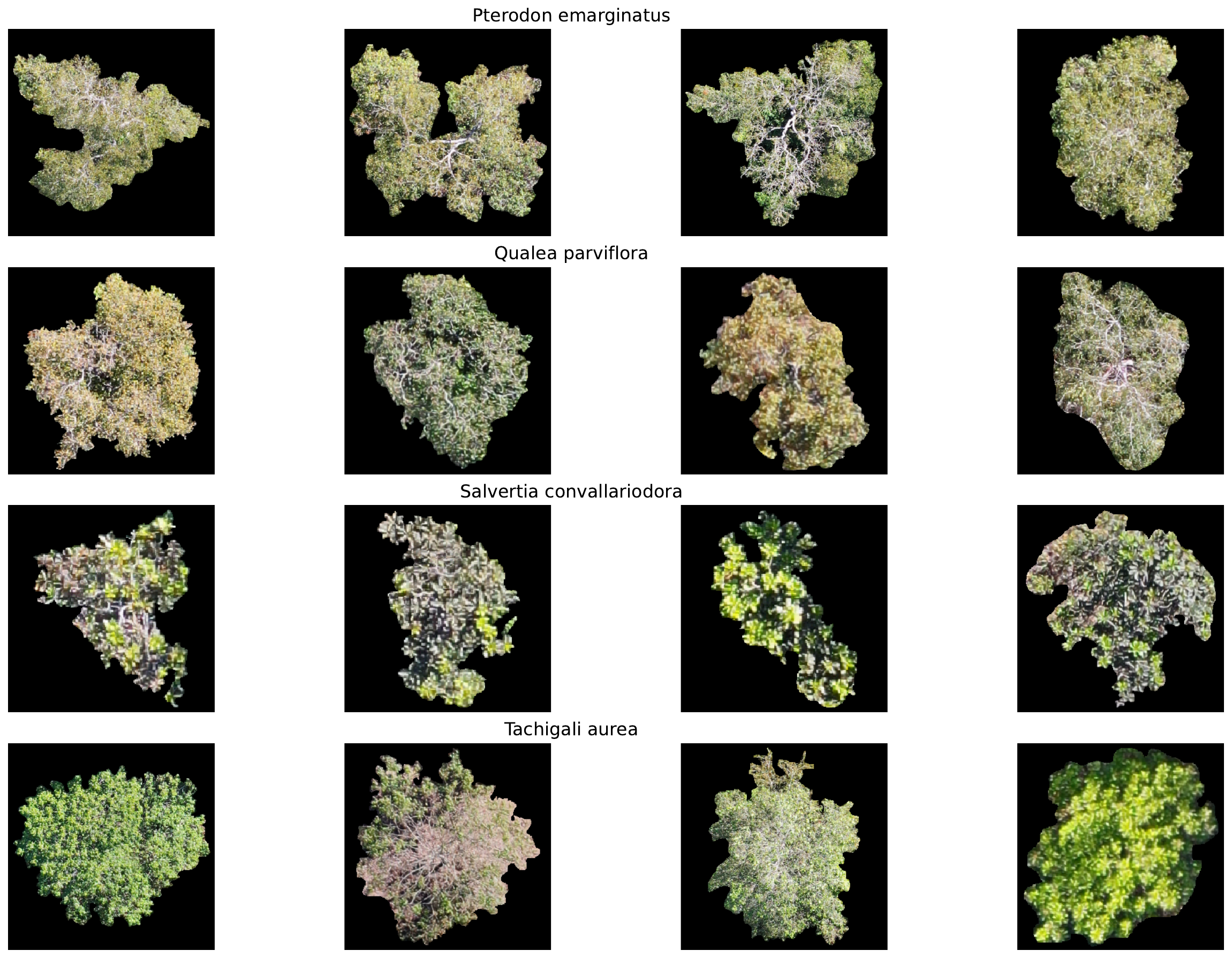}
	\caption{Examples of individual tree crowns of the four target species in the Bioflore MA dataset (UAV RGB imagery, 1.8 cm GSD).}
	\label{fig:figure_four_species_contrasting_crowns}
\end{figure*}

The second dataset used in this work comprises four species from three regions (see Figure \ref{fig:study_areas}) in localization (B). These species were selected based on two criteria: specific canopy characteristics (e.g., physical traits like shape, size, and color that enable visual differentiation among individuals) and their regional importance, encompassing either high economic value or an endangered status, thereby highlighting their significant financial and ecological relevance. Similar to the Amazon AC dataset, Figure \ref{fig:figure_four_species_contrasting_crowns} shows examples of UAV images illustrating the different species in the dataset. Table \ref{tab:tree_species_data_1} summarizes the species together with the corresponding distribution between training and testing sets.

\begin{table}[t]
    \setlength\tabcolsep{1.5pt}
    \centering
    \caption{Summary of ITCs for the tree species in Bioflore MA dataset}
    \begin{adjustbox}{width=\columnwidth,center}
    \begin{tabular}{c|c|c|c|c|c|c}
    \hline
         Index & Species name             & Train ITCs & Test ITCs & Total ITCs & Train ITCs (\%) & Test ITCs (\%)  \\
    \hline
         0     & Pterodon emarginatus     &         14 &         4 &         18 &          77.78	 &          22.22   \\
         1     & Qualea parviflora	     &         10 &     	3 &     	13 &          76.92  & 	        23.08   \\
         2     & Salvertia convallariodora&	       11 &	        2 &	        13 &	      84.62	 &          15.38   \\
         3     & Tachigali aurea	         &         16 &	        4 &	        20 &	      80.00	 &          20.00   \\
    \hline
    \end{tabular}
    \end{adjustbox}
    \label{tab:tree_species_data_1}
\end{table}

\subsection{CNN architecture}

The network architecture, conceptually depicted in Figure \ref{fig:multitask_architecture}, is a multi-task CNN designed for simultaneous semantic segmentation and distance transform. This DeepLabV3-based framework \cite{chen2017rethinking, cuerosa2021multi} employs a shared encoder network, built upon the ResNet-50 architecture \cite{he2016deep}, responsible for extracting hierarchical feature representations from the input image. The ResNet-50 structure is composed of 50 layers, including 16 residual blocks and 48 convolutional layers. As prescribed by the DeepLabV3 architecture \cite{chen2017rethinking}, the strided convolutions in the two deepest residual stages (layer3 and layer4) are replaced with atrous convolutions (dilation rates of 2 and 4, respectively), resulting in an output stride of 8 rather than the original 32. The encoder thus produces feature maps at 1/8 of the input spatial resolution, retaining finer spatial detail for pixel-level prediction. Pre-trained weights on the COCO Train 2017 dataset were used to initialize the network.

Branching from this shared encoder, two specialized decoder networks operate in parallel, as illustrated in Figure \ref{fig:multitask_architecture}. One decoder is dedicated to the main task of semantic segmentation, assigning a tree species label to each pixel. The other decoder handles the auxiliary task of distance transform, predicting the Euclidean distance from each pixel to the nearest ITC boundary. This multi-task design enables precise ITC delineation and classification by combining the discriminative power of segmentation with the boundary-aware regularization of the distance transform, proving suitable for our tree recruitment  strategy.

\subsection{Experimental setup}
\begin{figure*}[t]
  \centering
  \includegraphics[width=0.5\textwidth, angle=0]{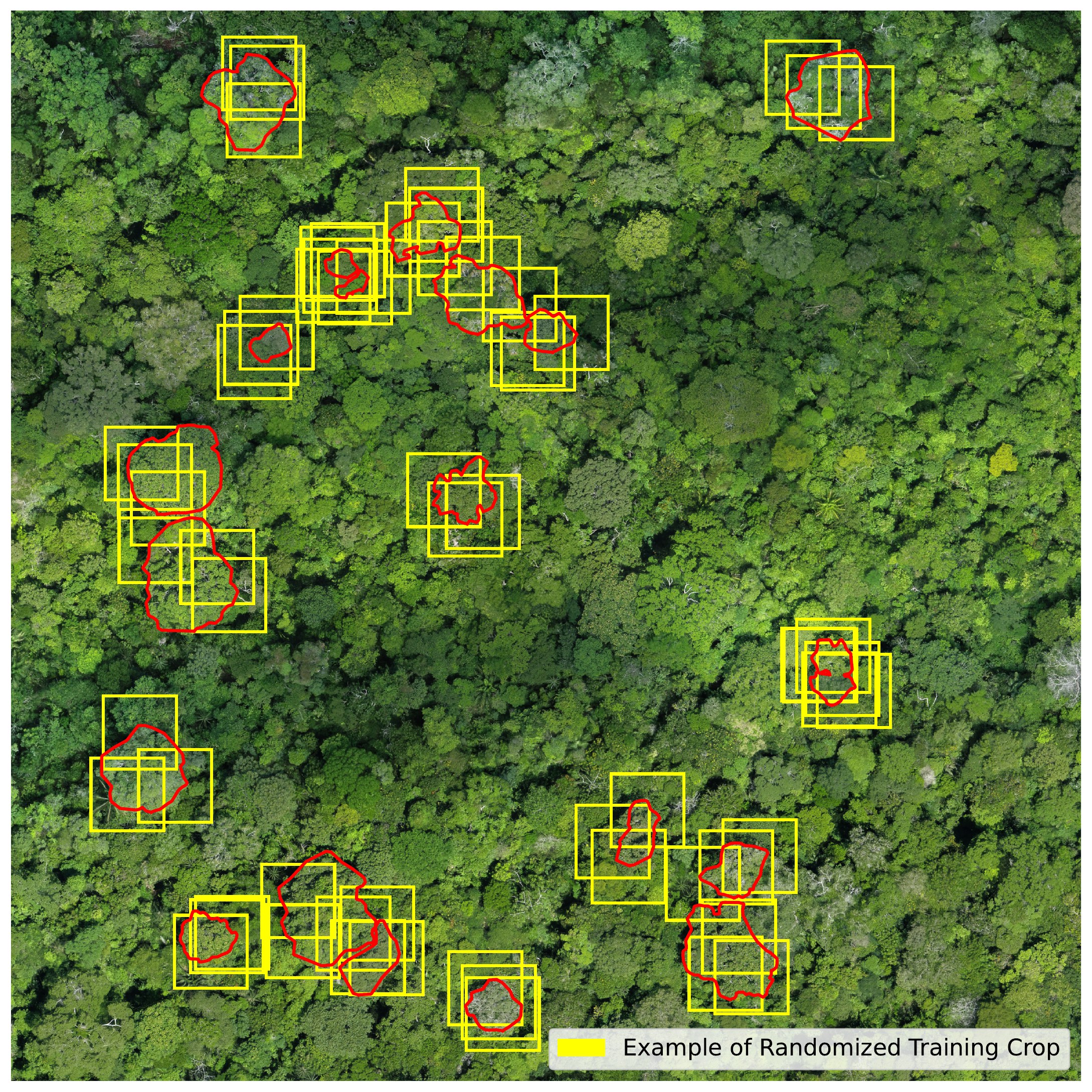}
  \caption{Examples of patches used in the training stage. Red polygons refer to individual tree crowns.}
  \label{fig:figure9}
\end{figure*}

In both datasets, the tree recruitment cycle started from the training ITCs listed in Tables \ref{tab:tree_species_data} and \ref{tab:tree_species_data_1} (355 crowns for Amazon AC and 51 crowns for Bioflore MA). To facilitate efficient processing and optimize memory consumption, images were processed as patches of $512 \times 512$ pixels for Amazon AC and $256 \times 256$ in Bioflore MA, with a batch size of 32. For patch selection, a pixel within an ITC was randomly chosen, and the patch dimensions were defined around this pixel. The distance from the selected pixel to the left edge and the top margin varied randomly between 1\% and 99\% of the patch's total dimension. This strategy, illustrated in Figure \ref{fig:figure9}, ensured consistent training with patches containing segmented pixels and presented the model with diverse positional contexts for the same sample.

Differences in the number of samples and segmented pixels were observed among tree species (as detailed in Tables \ref{tab:tree_species_data}, \ref{tab:tree_species_data_1}), leading to data imbalance that could potentially bias classification. To counteract this, an oversampling strategy was implemented to ensure a uniform representation of each tree species during training. Furthermore, to compensate for the limited annotated data, data augmentation techniques were applied randomly at each training epoch, including random rotations (90°, 180°, and 270°) and vertical and horizontal flips. To prevent overfitting, a dropout layer with a rate of 50\% was also integrated into the classification task. Additionally, in our experiments, the $\gamma$ parameter in equation \ref{equ:LPCFL} was set to 2 based on empirical tests and common practices in similar semantic segmentation tasks to address class imbalance.

The complete tree recruitment  cycle, depicted in Figure \ref{fig:model_overview}, was executed for 20 iterations. In each iteration, the model underwent training for a maximum of 30 epochs, with 5000 samples per epoch. The number of epochs was adjusted using an early stopping mechanism: training ceased if the F1-Score improvement on the validation set was less than 0.001 for five consecutive epochs. A validation set, constituting 20\% of the training set (ensuring at least one sample of each ITC per dataset), was separated at the start of each iteration's training. The network's optimization was performed using the stochastic gradient descent (SGD) optimizer with a momentum of 0.9, an initial learning rate of 0.01 with a linear warmup over 5 epochs and cosine annealing to 1e-4. The network was initialized with weights pre-trained on the COCO\footnote{https://cocodataset.org/\#home} Train 2017 dataset.

 Following \cite{cuerosa2021multi}, at inference time, we applied the trained network to overlapping image tiles using a sliding window strategy, averaging overlapping patch predictions pixel-wise, and improving results by considering different overlap rates. Hence, we generated the classification outputs for 30\% and 50\% overlap and took the average as the final outcome. In the same way as proposed in \cite{cuerosa2021multi}, we averaged the outputs to obtain an outcome with the input orthoimage dimensions.

All experiments were conducted on an Ubuntu 22.04 operating system, utilizing an AMD EPYC 7V13 CPU with 256GB of RAM, and an NVIDIA A100 graphics card with 80GB of VRAM. The entire model implementation was developed using the PyTorch framework.

\subsection{Performance Evaluation}

Model performance was evaluated on an independent test set, comprising samples explicitly excluded from the training phase. The F1-Score was designated as the primary metric for assessing performance, with values reported as averages across the testing ITCs. Specifically, for each individual class $c$, the F1-score is computed as the harmonic mean of Precision ($P_c$) and Recall ($R_c$), defined by:
\begin{equation}
F1-score_c = 2 \times \frac{P_c \times R_c}{P_c + R_c}
\label{eq:f1-score}
\end{equation}
where Precision ($P_c$) and Recall ($R_c$) are given by:
\begin{equation}
P_c = \frac{tp}{tp + fp}
\label{eq:precision}
\end{equation}
\begin{equation}
R_c = \frac{tp}{tp + fn}
\label{eq:recall}
\end{equation}
In these equations, $tp$ (true positives) represents the number of samples correctly assigned to class $c$. $fp$ (false positives) denotes the number of samples erroneously classified as class $c$, and $fn$ (false negatives) corresponds to the number of samples incorrectly classified as not belonging to class $c$. 

A consideration regarding the performance metrics stems from the nature of the remote sensing datasets used in this study. Unlike many semantic segmentation tasks that often benefit from exhaustive, pixel-level annotations across the entire image, our data does not provide this complete level of detail. Instead, we possess a limited set of annotated ITCs, represented as polygons within the image space, specifically for the tree species of interest.

This characteristic directly constrains our evaluation methodology, making the computation of traditional segmentation metrics, such as pixel-level Precision and Recall, unfeasible across the entire image. To ensure an assessment of the model's performance, we calculated metrics at the ITC level. Each annotated ITC was treated as an individual instance, and the model's evaluation focused on its correct detection and classification, similar to an image classification task where each object is assessed individually. The final class prediction was determined by the majority class within the polygon region. This approach quantifies the model's ability to accurately identify existing ITCs and assign them to the correct class, thereby aligning the evaluation with both the characteristics of the available data and the objectives of the task.

\FloatBarrier
\section{Results and Discussion}
\label{Results}
In this section, we present the results of the experiments conducted in this study. We start by comparing the performances of DL-based models already proposed in the literature, e.g. DeepLabV3 \cite{chen2017rethinking}, and the Multi-Task method proposed in \cite{cuerosa2021multi} against the results obtained by the proposed method on each site. Next, we present results obtained by each specie and visually analyze the classification difficulties derived from the set of individual predictions of the evaluated methods.

\subsection{Instance Tree Crowns Segmentation Accuracy}

Table \ref{tab:results} presents the performance of the evaluated segmentation methods, namely DeepLabV3 \cite{chen2017rethinking}, a Multi-task DeepLabV3 \cite{cuerosa2021multi}, and our proposed approach. The performance is quantified by the F1-score, Precision, and Recall metrics, averaged across all categories within the Amazon AC and BioFlore MA datasets, respectively.  The best results obtained for each metric are highlighted in bold.

\begin{table}[t]
\centering
\setlength\tabcolsep{1.5pt}
\caption{Average F1-score, Precision and Recall (\%) of each method for both datasets.}
\label{tab:results}
\begin{adjustbox}{width=\columnwidth,center}
\begin{tabular}{c|ccc|ccc}
\hline
Datasets  & \multicolumn{3}{c|}{Amazon AC}                                    & \multicolumn{3}{c}{BioFlore MA}                                        \\ \hline
Metrics   & \multicolumn{1}{c|}{F1-score} & \multicolumn{1}{c|}{Precision} & Recall & \multicolumn{1}{c|}{F1-score} & \multicolumn{1}{c|}{Precision} & Recall \\ \hline
DeepLabV3 \cite{chen2017rethinking} & \multicolumn{1}{c|}{50.0}     & \multicolumn{1}{c|}{53.1}      & 51.6   & \multicolumn{1}{c|}{79.7}     & \multicolumn{1}{c|}{81.6}      & 79.1   \\
Multi-task \cite{cuerosa2021multi}    & \multicolumn{1}{c|}{45.2}     & \multicolumn{1}{c|}{47.3}      & 47.6   & \multicolumn{1}{c|}{83.6}     & \multicolumn{1}{c|}{86.6}      & 85.4   \\
Automatic Tree Recruitment -MT (Ours)      & \multicolumn{1}{c|}{\textbf{53.5}}     & \multicolumn{1}{c|}{\textbf{55.9}}      & \textbf{54.6}   & \multicolumn{1}{c|}{\textbf{92.2}}     & \multicolumn{1}{c|}{\textbf{95.0}}      & \textbf{91.6}   \\ \hline
\end{tabular}
\end{adjustbox}
\end{table}

It can be observed by inspecting Table \ref{tab:results} that the proposed method is consistently superior across all evaluated metrics ,i.e., F1-score, Precision, and Recall, for both the Amazon AC and BioFlore MA datasets. Specifically, in the Amazon AC dataset, our approach achieved the highest scores with 53.5\% F1-score, 55.9\% Precision, and 54.6\% Recall. Similarly, it outperformed all other methods in the BioFlore MA dataset, registering 92.2\% F1-score, 95.0\% Precision, and 91.6\% Recall.

A general trend observed across all methods is a higher performance on the BioFlore MA dataset compared to Amazon AC. This evidences the differences with respect to characteristics such as inter-class variability, clearer object boundaries (see Figures \ref{fig:figure_seventeen_species_contrasting_crowns} and \ref{fig:figure_four_species_contrasting_crowns}) that make the Amazon AC dataset more challenging. Striking is the performance of the multi-task baseline \cite{cuerosa2021multi} on the Amazon AC dataset, where it recorded an F1-score of 45.2\%, along with a Precision (47.3\%) and Recall (47.6\%). This result indicates a limitation on the Amazon AC dataset, a point that warrants investigation in the subsequent steps of this work.

While DeepLabV3 served as a baseline, demonstrating competitive results in BioFlore MA with an F1-score of 79.7\%, it was consistently outperformed by the proposed method and the multi-task proposal from \cite{cuerosa2021multi}. This superiority, especially when contrasted with the performance obtained by the established multi-task baseline, highlights the benefits derived by including the proposed tree recruitment  strategy. These results underscore the potential of our contributions in advancing multi-task learning for challenging environmental remote sensing data.

\begin{figure*}[t]
    \centering
    \begin{subfigure}[b]{0.85\textwidth}
        \centering
         \includegraphics[width=\textwidth]{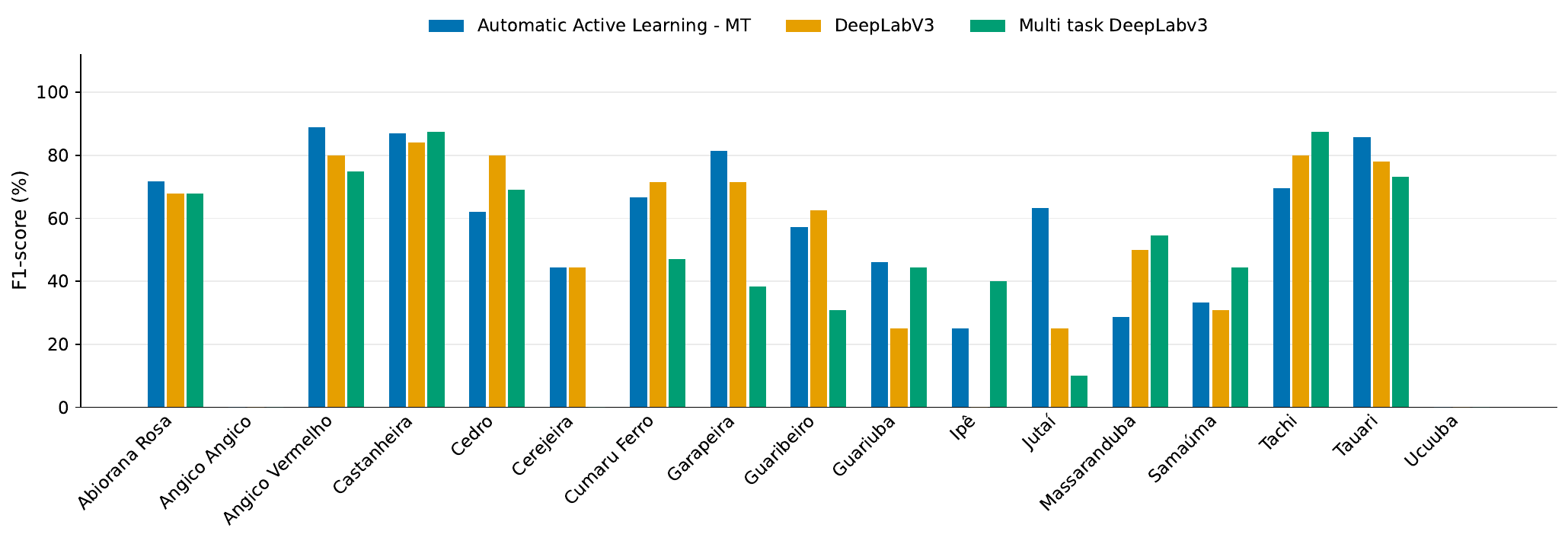}
         \caption{Amazon AC}
         \label{fig:barplot_amazon_ac}
    \end{subfigure}
    \hfill
    \begin{subfigure}[b]{0.65\textwidth}
        \centering
         \includegraphics[width=\textwidth]{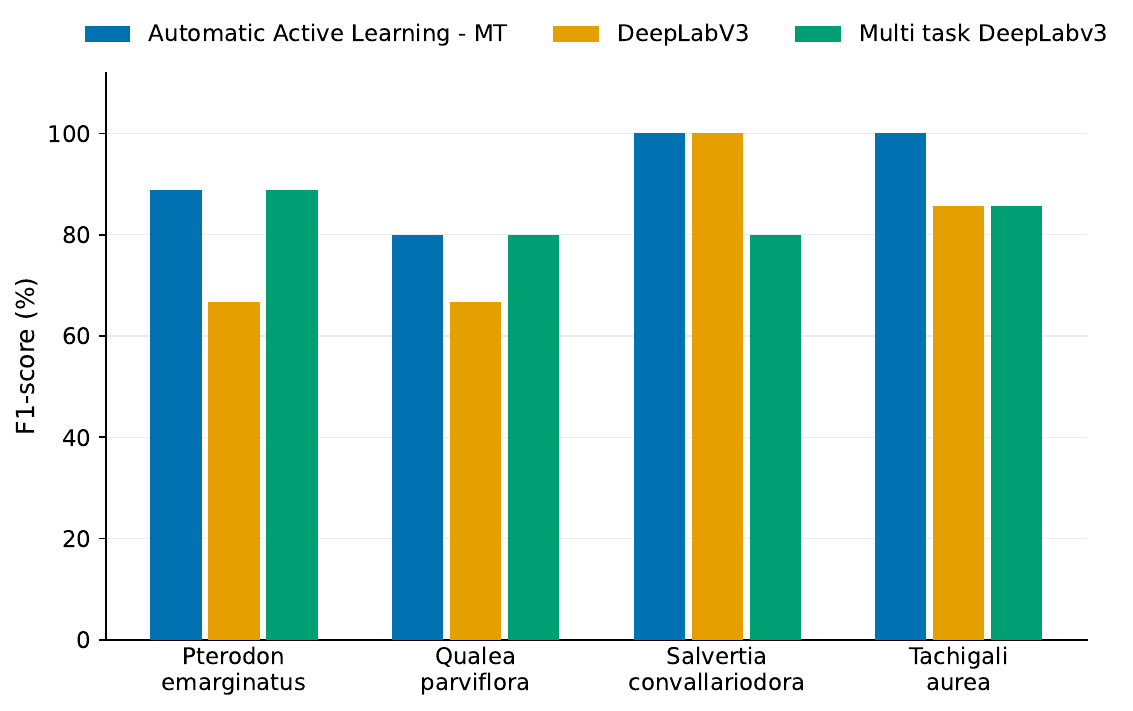}
         \caption{Bioflore MA}
         \label{fig:barplot_bioflore_ma}
    \end{subfigure}
    \caption{F1-score (\%) for each specie in both datasets}
    \label{fig:barplots}
\end{figure*}

To provide a deeper, species-level understanding of the previously analyzed overall results, Figures \ref{fig:barplot_amazon_ac} and \ref{fig:barplot_bioflore_ma} present the detailed F1-score (\%) obtained by each evaluated method for individual species within the Amazon AC and BioFlore MA datasets, respectively. In these figures, each cluster of bars represents the performance for a specific species. The green bars correspond to the Multi-task baseline proposed by \cite{cuerosa2021multi}, the yellow bars illustrate the DeepLabV3 \cite{chen2017rethinking} results, and the blue bars depict the performance achieved with the proposed method. 

As corroborated by the overall findings in Table \ref{tab:results}, all evaluated methods demonstrated performance on the BioFlore MA dataset, with F1-scores generally surpassing 60\%, and for the leading approaches, frequently exceeding 80\%, see Figure \ref{fig:barplot_bioflore_ma}. Within this dataset, \textit{Qualea parviflora} presented the most significant segmentation challenge, yielding the lowest F1-scores across all models. A potential explanation for this difficulty may stem from the nature of its canopy structure, which, despite its complexity, might share subtle morphological characteristics with \textit{Pterodon emarginatus} species present in the dataset, see Figure \ref{fig:figure_four_species_contrasting_crowns}. This inter-species resemblance could lead to increased ambiguity for the models, hindering precise differentiation and subsequently impacting segmentation accuracy for \textit{Qualea parviflora}. Conversely, higher performances were observed for species such as \textit{Salvertia convallariodora} and \textit{Tachigali aurea} where models achieved F1-scores between 80\% and 100\%. The proposed method outperformed DeepLabV3 across most species in this dataset, while achieving comparable F1-scores to the Multi-Task DeepLabV3 method for \textit{Pterodon emarginatus} and \textit{Qualea parviflora}. We hypothesize that the varying performance across species may be correlated with the complexity of their canopy shape and the indistinctness of their boundaries. This is a result that deserves further experimentation, potentially involving a greater number of samples or more refined morphological feature extraction.

In contrast, an analysis of Figure \ref{fig:barplot_amazon_ac}, which illustrates the comparative performance of the methods for Amazon AC species, evidences the complexity of the task, as demonstrated by the variability in the results achieved by each approach. Consequently, Figure \ref{fig:barplot_amazon_ac} highlights \textit{Angico} and \textit{Ucuuba} as the most challenging species for accurate detection. Specifically, for \textit{Angico} and \textit{Ucuuba}, no method demonstrated the capability to detect their instances. This challenge is likely attributable to their underrepresentation within the dataset inventory, characterized by a considerably lower number of delineated crowns compared to other included species, see Table \ref{tab:tree_species_data}. Furthermore, the methods yielded relatively low accuracies, specifically ranging from 10\% to 65\%, for species such as \textit{Cerejeira, Guariuba, Ipê, Jutai, Massaranduba}, and \textit{Samauma}. Although the training sample sizes for these species were slightly larger than those for the challenging cases previously discussed, they still represent a limitation for proper training and learning. Furthermore, similar to observations in the Bioflore MA dataset, canopy structures within each species exhibit variability across individual instances. This diversity renders visual analysis challenging, even for expert human observers. Figure \ref{fig:figure_seventeen_species_contrasting_crowns} illustrates this phenomenon, highlighting the distinct canopy shape and texture for instances of \textit{Ipê, Guariuba, Cerejeira}, and \textit{Jutaí}. Notwithstanding the aforementioned challenges, the proposed method exhibited superior performance over both the Single and Multi-Task versions of DeepLabV3 across most species, while achieving lower or comparable results with these methods in the remaining species.

\begin{figure*}[t]
    \centering
    \begin{tikzpicture}[
            imgenode/.style={rectangle,  minimum width = 0.5cm, minimum height = 0.5 cm},
            textnode/.style={rectangle,  minimum width = 0.01cm, minimum height = 0.01 cm, text centered},
            circlenode/.style={circle, draw = black,  inner sep=0pt, minimum size=4pt, fill = white},
            ]
            \node(AR_0)[imgenode] {\includegraphics[width=.215\textwidth]{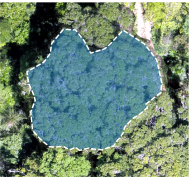}};
            \node(AR_1)[imgenode, right=-0.15cm of AR_0] {\includegraphics[width=.215\textwidth]{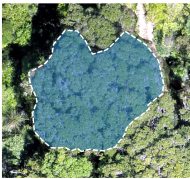}};
            \node(AR_2)[imgenode, right=-0.15cm of AR_1] {\includegraphics[width=.215\textwidth]{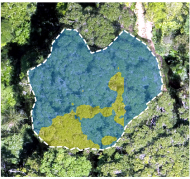}};
            \node(AR_3)[imgenode, right=-0.15cm of AR_2] {\includegraphics[width=.215\textwidth]{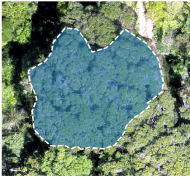}};

            \node(AV_0)[imgenode, below= 0.65cm of AR_0] {\includegraphics[width=.215\textwidth]{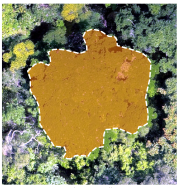}};
            \node(AV_1)[imgenode, right=-0.15cm of AV_0] {\includegraphics[width=.215\textwidth]{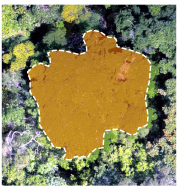}};
            \node(AV_2)[imgenode, right=-0.15cm of AV_1] {\includegraphics[width=.215\textwidth]{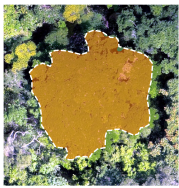}};
            \node(AV_3)[imgenode, right=-0.15cm of AV_2] {\includegraphics[width=.215\textwidth]{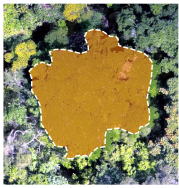}};

            \node(SA_0)[imgenode, below= 0.65cm of AV_0] {\includegraphics[width=.215\textwidth]{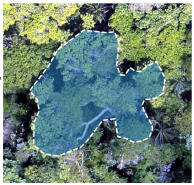}};
            \node(SA_1)[imgenode, right=-0.15cm of SA_0] {\includegraphics[width=.215\textwidth]{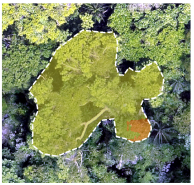}};
            \node(SA_2)[imgenode, right=-0.15cm of SA_1] {\includegraphics[width=.215\textwidth]{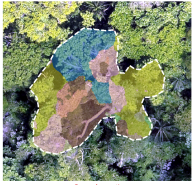}};
            \node(SA_3)[imgenode, right=-0.15cm of SA_2] {\includegraphics[width=.215\textwidth]{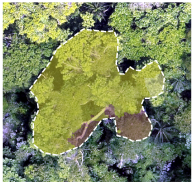}};

            \node(AN_0)[imgenode, below= 0.65cm of SA_0] {\includegraphics[width=.215\textwidth]{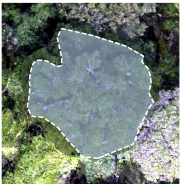}};
            \node(AN_1)[imgenode, right=-0.15cm of AN_0] {\includegraphics[width=.215\textwidth]{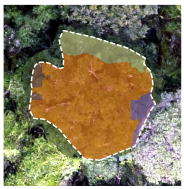}};
            \node(AN_2)[imgenode, right=-0.15cm of AN_1] {\includegraphics[width=.215\textwidth]{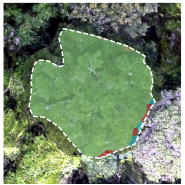}};
            \node(AN_3)[imgenode, right=-0.15cm of AN_2] {\includegraphics[width=.215\textwidth]{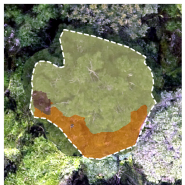}};

            \node(title_AR_0)[textnode, above = -0.25 cm of AR_0] {\small{Ground-Truth}};
            \node(title_AR_1)[textnode, above = -0.25 cm of AR_1] {\small{DeepLabV3}};
            \node(title_AR_2)[textnode, above = -0.25 cm of AR_2] {\small{\begin{tabular}{c} Multi-Task\\ DeepLabV3 \end{tabular}}};
            \node(title_AR_3)[textnode, above = -0.25 cm of AR_3] {\small{\begin{tabular}{c} Automatic Active \\ Learning-MT \end{tabular}}};

            \node(leg11_1)[textnode, below = -0.2 cm of AR_0]{\footnotesize{(a)}};
            \node(leg11_2)[textnode, below = -0.2 cm of AR_1]{\footnotesize{(b)}};
            \node(leg11_3)[textnode, below = -0.2 cm of AR_2]{\footnotesize{(c)}};
            \node(leg11_4)[textnode, below = -0.2 cm of AR_3]{\footnotesize{(d)}};

            \node(leg11_1_1)[textnode, below = 0.1 cm of AR_0]{\footnotesize{\textbf{TL}: Abiorana rosa}};
            \node(leg11_2_1)[textnode, below = 0.1 cm of AR_1]{\footnotesize{\textbf{PL}: Abiorana rosa}};
            \node(leg11_3_1)[textnode, below = 0.1 cm of AR_2]{\footnotesize{\textbf{PL}: Abiorana rosa}};
            \node(leg11_4_1)[textnode, below = 0.1 cm of AR_3]{\footnotesize{\textbf{PL}: Abiorana rosa}};

            \node(leg12_1)[textnode, below = -0.2 cm of AV_0]{\footnotesize{(e)}};
            \node(leg12_2)[textnode, below = -0.2 cm of AV_1]{\footnotesize{(f)}};
            \node(leg12_3)[textnode, below = -0.2 cm of AV_2]{\footnotesize{(g)}};
            \node(leg12_4)[textnode, below = -0.2 cm of AV_3]{\footnotesize{(h)}};

            \node(leg12_1_1)[textnode, below = 0.1 cm of AV_0]{\footnotesize{\textbf{TL}: Angico vermelho}};
            \node(leg12_2_1)[textnode, below = 0.1 cm of AV_1]{\footnotesize{\textbf{PL}: Angico vermelho}};
            \node(leg12_3_1)[textnode, below = 0.1 cm of AV_2]{\footnotesize{\textbf{PL}: Angico vermelho}};
            \node(leg12_4_1)[textnode, below = 0.1 cm of AV_3]{\footnotesize{\textbf{PL}: Angico vermelho}};

            \node(leg13_1)[textnode, below = -0.2 cm of SA_0]{\footnotesize{(i)}};
            \node(leg13_2)[textnode, below = -0.2 cm of SA_1]{\footnotesize{(j)}};
            \node(leg13_3)[textnode, below = -0.2 cm of SA_2]{\footnotesize{(k)}};
            \node(leg13_4)[textnode, below = -0.2 cm of SA_3]{\footnotesize{(l)}};

            \node(leg13_1_1)[textnode, below = 0.1 cm of SA_0]{\footnotesize{\textbf{TL}: Abiorana rosa}};
            \node(leg13_2_1)[textnode, below = 0.1 cm of SA_1]{\footnotesize{\textbf{PL}: Samauma}};
            \node(leg13_3_1)[textnode, below = 0.1 cm of SA_2]{\footnotesize{\textbf{PL}: Samauma}};
            \node(leg13_4_1)[textnode, below = 0.1 cm of SA_3]{\footnotesize{\textbf{PL}: Samauma}};

            \node(leg14_1)[textnode, below = -0.2 cm of AN_0]{\footnotesize{(m)}};
            \node(leg14_2)[textnode, below = -0.2 cm of AN_1]{\footnotesize{(n)}};
            \node(leg14_3)[textnode, below = -0.2 cm of AN_2]{\footnotesize{(o)}};
            \node(leg14_4)[textnode, below = -0.2 cm of AN_3]{\footnotesize{(p)}};

            \node(leg14_1_1)[textnode, below = 0.1 cm of AN_0]{\footnotesize{\textbf{TL}: Angico}};
            \node(leg14_2_1)[textnode, below = 0.1 cm of AN_1]{\footnotesize{\textbf{PL}: Angico vermelho}};
            \node(leg14_3_1)[textnode, below = 0.1 cm of AN_2]{\footnotesize{\textbf{PL}: Cedro}};
            \node(leg14_4_1)[textnode, below = 0.1 cm of AN_3]{\footnotesize{\textbf{PL}: Tachi}};
    \end{tikzpicture}
    \caption{Visual analysis of predicted ITCs by each evaluated method on Amazon AC dataset. In the figure, \textbf{TL}, and \textbf{PL} refer to True Label and Predicted Label, respectively.}
    \label{fig:amzon_ac_vis}
\end{figure*}

\begin{figure*}[t]
    \centering
    \begin{tikzpicture}[
            imgenode/.style={rectangle,  minimum width = 0.5cm, minimum height = 0.5 cm},
            textnode/.style={rectangle,  minimum width = 0.01cm, minimum height = 0.01 cm, text centered},
            circlenode/.style={circle, draw = black,  inner sep=0pt, minimum size=4pt, fill = white},
            ]
            \node(AR_0)[imgenode] {\includegraphics[width=.215\textwidth]{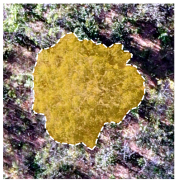}};
            \node(AR_1)[imgenode, right=-0.15cm of AR_0] {\includegraphics[width=.215\textwidth]{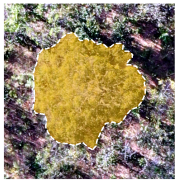}};
            \node(AR_2)[imgenode, right=-0.15cm of AR_1] {\includegraphics[width=.215\textwidth]{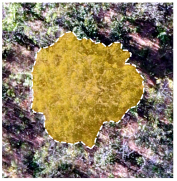}};
            \node(AR_3)[imgenode, right=-0.15cm of AR_2] {\includegraphics[width=.215\textwidth]{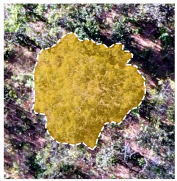}};

            \node(AV_0)[imgenode, below= 0.65cm of AR_0] {\includegraphics[width=.215\textwidth]{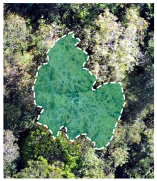}};
            \node(AV_1)[imgenode, right=-0.15cm of AV_0] {\includegraphics[width=.215\textwidth]{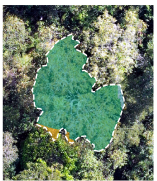}};
            \node(AV_2)[imgenode, right=-0.15cm of AV_1] {\includegraphics[width=.215\textwidth]{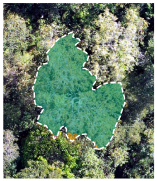}};
            \node(AV_3)[imgenode, right=-0.15cm of AV_2] {\includegraphics[width=.215\textwidth]{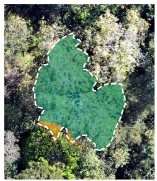}};

            \node(SA_0)[imgenode, below= 0.65cm of AV_0] {\includegraphics[width=.215\textwidth]{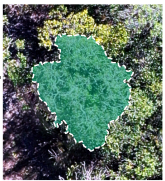}};
            \node(SA_1)[imgenode, right=-0.15cm of SA_0] {\includegraphics[width=.215\textwidth]{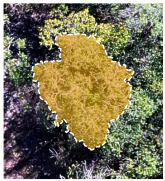}};
            \node(SA_2)[imgenode, right=-0.15cm of SA_1] {\includegraphics[width=.215\textwidth]{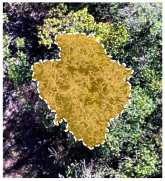}};
            \node(SA_3)[imgenode, right=-0.15cm of SA_2] {\includegraphics[width=.215\textwidth]{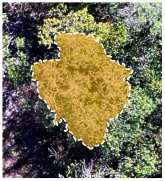}};

            \node(title_AR_0)[textnode, above = -0.25 cm of AR_0] {\small{Ground-Truth}};
            \node(title_AR_1)[textnode, above = -0.25 cm of AR_1] {\small{DeepLabV3}};
            \node(title_AR_2)[textnode, above = -0.25 cm of AR_2] {\small{\begin{tabular}{c} Multi-Task\\ DeepLabV3 \end{tabular}}};
            \node(title_AR_3)[textnode, above = -0.25 cm of AR_3] {\small{\begin{tabular}{c} Automatic Active \\ Learning-MT \end{tabular}}};

            \node(leg11_1)[textnode, below = -0.2 cm of AR_0]{\footnotesize{(a)}};
            \node(leg11_2)[textnode, below = -0.2 cm of AR_1]{\footnotesize{(b)}};
            \node(leg11_3)[textnode, below = -0.2 cm of AR_2]{\footnotesize{(c)}};
            \node(leg11_4)[textnode, below = -0.2 cm of AR_3]{\footnotesize{(d)}};

            \node(leg11_1_1)[textnode, below = 0.1 cm of AR_0]{\scriptsize{\textbf{TL}: Pterodon emarginatus}};
            \node(leg11_2_1)[textnode, below = 0.1 cm of AR_1]{\scriptsize{\textbf{PL}: Pterodon emarginatus}};
            \node(leg11_3_1)[textnode, below = 0.1 cm of AR_2]{\scriptsize{\textbf{PL}: Pterodon emarginatus}};
            \node(leg11_4_1)[textnode, below = 0.1 cm of AR_3]{\scriptsize{\textbf{PL}: Pterodon emarginatus}};

            \node(leg12_1)[textnode, below = -0.2 cm of AV_0]{\footnotesize{(e)}};
            \node(leg12_2)[textnode, below = -0.2 cm of AV_1]{\footnotesize{(f)}};
            \node(leg12_3)[textnode, below = -0.2 cm of AV_2]{\footnotesize{(g)}};
            \node(leg12_4)[textnode, below = -0.2 cm of AV_3]{\footnotesize{(h)}};

            \node(leg12_1_1)[textnode, below = 0.1 cm of AV_0]{\scriptsize{\textbf{TL}: Qualea parviflora}};
            \node(leg12_2_1)[textnode, below = 0.1 cm of AV_1]{\scriptsize{\textbf{PL}: Qualea parviflora}};
            \node(leg12_3_1)[textnode, below = 0.1 cm of AV_2]{\scriptsize{\textbf{PL}: Qualea parviflora}};
            \node(leg12_4_1)[textnode, below = 0.1 cm of AV_3]{\scriptsize{\textbf{PL}: Qualea parviflora}};

            \node(leg13_1)[textnode, below = -0.2 cm of SA_0]{\footnotesize{(i)}};
            \node(leg13_2)[textnode, below = -0.2 cm of SA_1]{\footnotesize{(j)}};
            \node(leg13_3)[textnode, below = -0.2 cm of SA_2]{\footnotesize{(k)}};
            \node(leg13_4)[textnode, below = -0.2 cm of SA_3]{\footnotesize{(l)}};

            \node(leg13_1_1)[textnode, below = 0.1 cm of SA_0]{\scriptsize{\textbf{TL}: Qualea parviflora}};
            \node(leg13_2_1)[textnode, below = 0.1 cm of SA_1]{\scriptsize{\textbf{PL}: Pterodon emarginatus}};
            \node(leg13_3_1)[textnode, below = 0.1 cm of SA_2]{\scriptsize{\textbf{PL}: Pterodon emarginatus}};
            \node(leg13_4_1)[textnode, below = 0.1 cm of SA_3]{\scriptsize{\textbf{PL}: Pterodon emarginatus}};
    \end{tikzpicture}
    \caption{Visual analysis of predicted ITCs by each evaluated method on Bioflore MA dataset. In the figure, \textbf{TL}, and \textbf{PL} refer to True Label and Predicted Label, respectively.}
    \label{fig:bioflore_vis}
\end{figure*}

We conclude the discussion with a qualitative analysis of the model predictions. Figures \ref{fig:amzon_ac_vis} and \ref{fig:bioflore_vis} show  subsets of prediction overlapped to the real RGB images from both Amazon AC and Bioflore MA respectively. The image subsets on the first column represents the respective ground-truth (Figures \ref{fig:amzon_ac_vis}(a)(e)(i)(m) and  \ref{fig:bioflore_vis}(a)(e)(i)). The subsets shown in the following columns show the outcome of the image predictions obtained from the different models: DeepLabV3 (Figures \ref{fig:amzon_ac_vis}(b)(f)(j)(n) and \ref{fig:bioflore_vis}(b)(f)(j)); Multi-Task DeepLabV3 (\ref{fig:amzon_ac_vis}(c)(g)(k)(o) and \ref{fig:bioflore_vis}(c)(g)(k)); and Automatic Tree Recruitment -MT (Figures \ref{fig:amzon_ac_vis}(d)(h)(l)(p) and \ref{fig:bioflore_vis}(d)(h)(l)). Each row corresponds to one sample and \textbf{TL} and \textbf{PL} denote \textbf{True Label} and \textbf{Predicted Label} respectively. Finally, ITCs are presented in all cases from the manual drawing provided in the ground-truth inventory.

Inspecting the images in Figure \ref{fig:amzon_ac_vis} and \ref{fig:bioflore_vis}, a primary observation is the spatial consistency of predicted species. Baseline models, the standard DeepLabV3 and the Multi-Task DeepLabV3 exhibit reasonable performance on some instances (Figures \ref{fig:amzon_ac_vis}(b)(f)(g) and \ref{fig:bioflore_vis}(b)(c)), while class fragmentation in others within individual tree canopies. As seen in Figures \ref{fig:amzon_ac_vis}(c)(j)(k)(n)(o) and \ref{fig:bioflore_vis}(f)(g), these models often predict multiple, disjointed species labels for a single continuous canopy area.

In contrast, the Automatic Tree Recruitment -MT approach (Figures \ref{fig:amzon_ac_vis}(d)(h)(l)(p) and \ref{fig:bioflore_vis}(d)(h)(l)) produces more homogeneous segmentation masks. By selecting informative samples during training, the model learns more robust feature representations that minimize spurious class transitions within a single object boundary.

\FloatBarrier
\subsection{Instance Tree Crowns Segmentation Complementary Results}

\begin{figure*}[t] \centering \includegraphics[width=1\textwidth, angle=0]{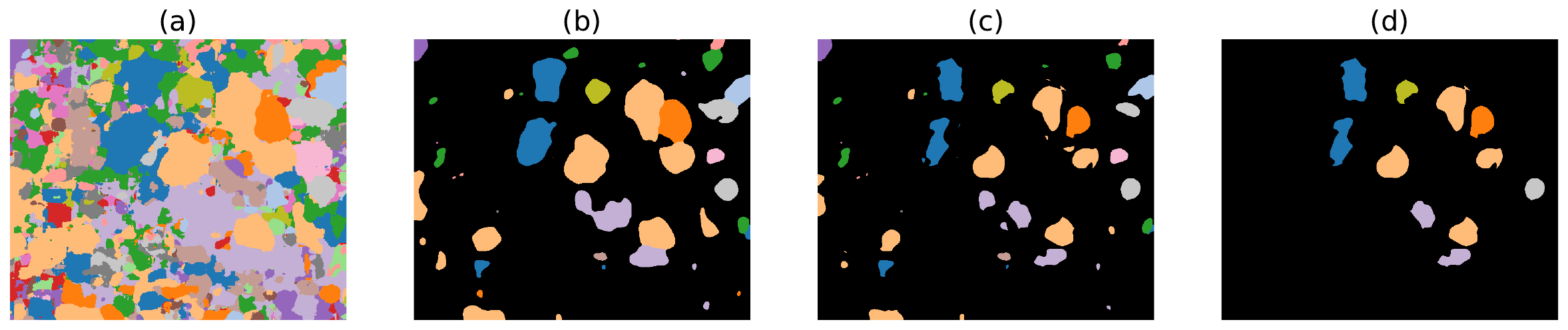} \caption{ Illustration of component selection based on model predictions: (a) Predicted semantic segmentation by the model. (b) Prediction with confidence above 70\%. (c) High-confidence prediction where the distance map value is above 0.25. (d) ITCs are selected to compose the tree recruitment  scheme sample inventory. } \label{fig:mapa_distancia_prob_classe} \end{figure*}

To better understand the operational mechanics of the proposed tree recruitment  method, especially regarding the computation of predictions and subsequent metrics, the process of ITC instance segmentation is visualized through a series of filtering steps, as illustrated in Figure \ref{fig:mapa_distancia_prob_classe}. This sequence, designed to enhance accuracy, begins with the model producing a semantically segmented image, where each pixel is assigned a label corresponding to a specific tree species (Figure \ref{fig:mapa_distancia_prob_classe}a). Applying a threshold to the probability image removes pixels with low confidence, thereby reducing the number of classified pixels and retaining only those with high-class membership (Figure \ref{fig:mapa_distancia_prob_classe}b). Further refinement occurs via a threshold on the distance transform image, improving ITC delineation (Figure \ref{fig:mapa_distancia_prob_classe}c). The final step involves selecting valid samples based on crown morphology, which enhances the precision of tree crown detection and contributes to the tree recruitment  sample inventory (Figure \ref{fig:mapa_distancia_prob_classe}d).

\begin{figure*}[t] \centering \includegraphics[width=1\textwidth]{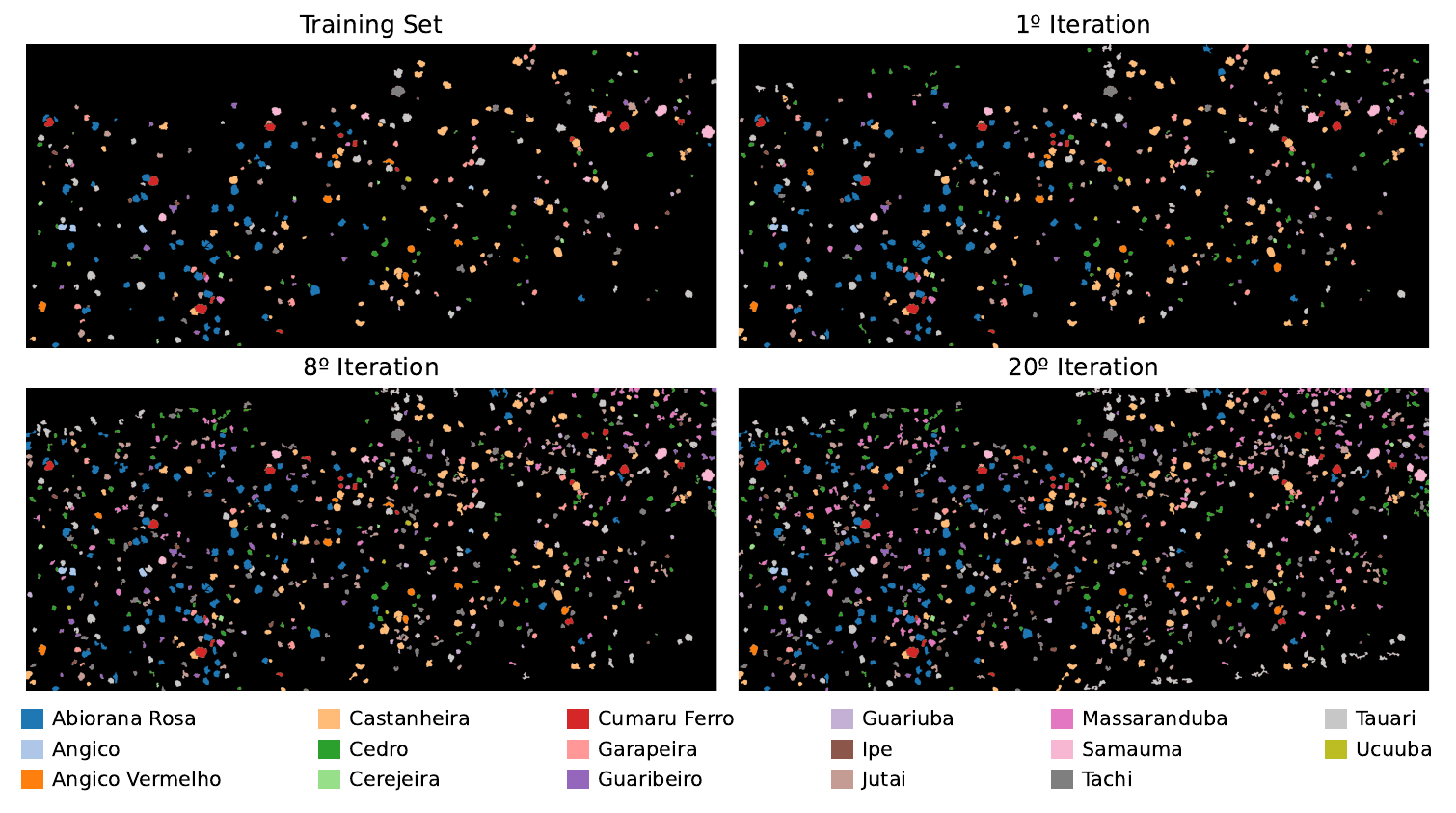} \caption{Individual tree crowns (ITCs) of the target species (see Table \ref{tab:tree_species_data}) over the iterations of the tree recruitment  process for Amazon AC dataset.} \label{fig:segmentation_evolution} \end{figure*}

As consequence, the proposed method increases the number of ITCs compared to the initial training set, which can be observed in Figure \ref{fig:segmentation_evolution} in different moments during algorithm's training. This expansion indicates a progressive improvement in the model's ability to identify and segment tree instances across tree recruitment  iterations.

\begin{figure}[t]
    \centering
    \includegraphics[width=0.7\linewidth, angle=0]{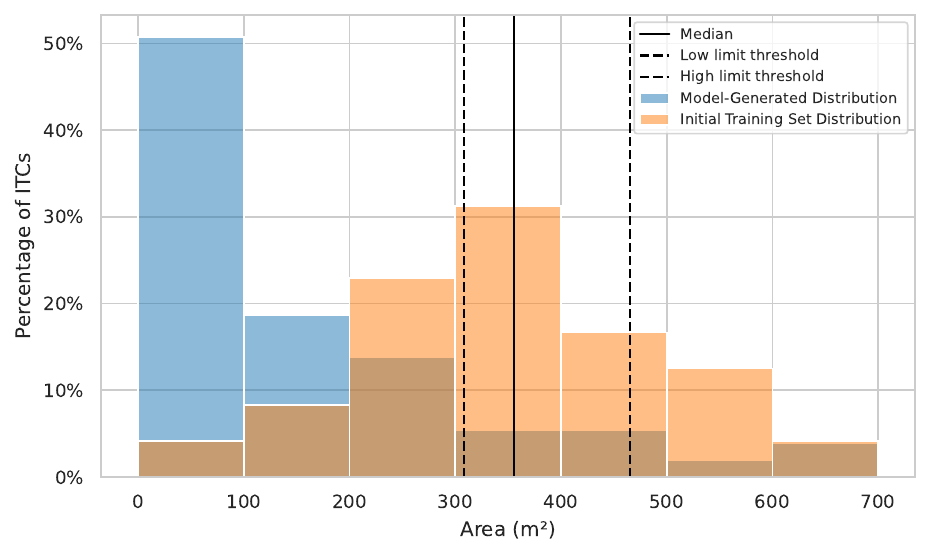}
    \caption{Comparison of the canopy area distribution of the species \textit{Abiorana rosa} (\textit{Micropholis venulosa}). Highlighted are the selection limits used in the first iteration of the model.}
    \label{fig:distribuicao_abiorana_rosa}
\end{figure}

Comparing the predicted crown area distributions with the reference training set is crucial for evaluating model performance. This comparison helps verify whether the model’s predictions remain aligned with the reference ITCs, ensuring that the model does not diverge from the actual crown areas. As shown in Figure \ref{fig:distribuicao_abiorana_rosa}, the distribution of crown areas for \textit{Abiorana rosa} (\textit{Micropholis venulosa}) is analyzed, using selection limits from the first iteration. Blue bars represent the predicted samples, reflecting the model's estimation of crown area distribution, while orange bars indicate the reference training ITCs. The results reveal that most predicted ITCs have areas ranging from 0 to 100 m², which are smaller than the median and consequently excluded from the sample inventory.

\begin{figure*}[t]
    \centering
    \includegraphics[width=1\textwidth]{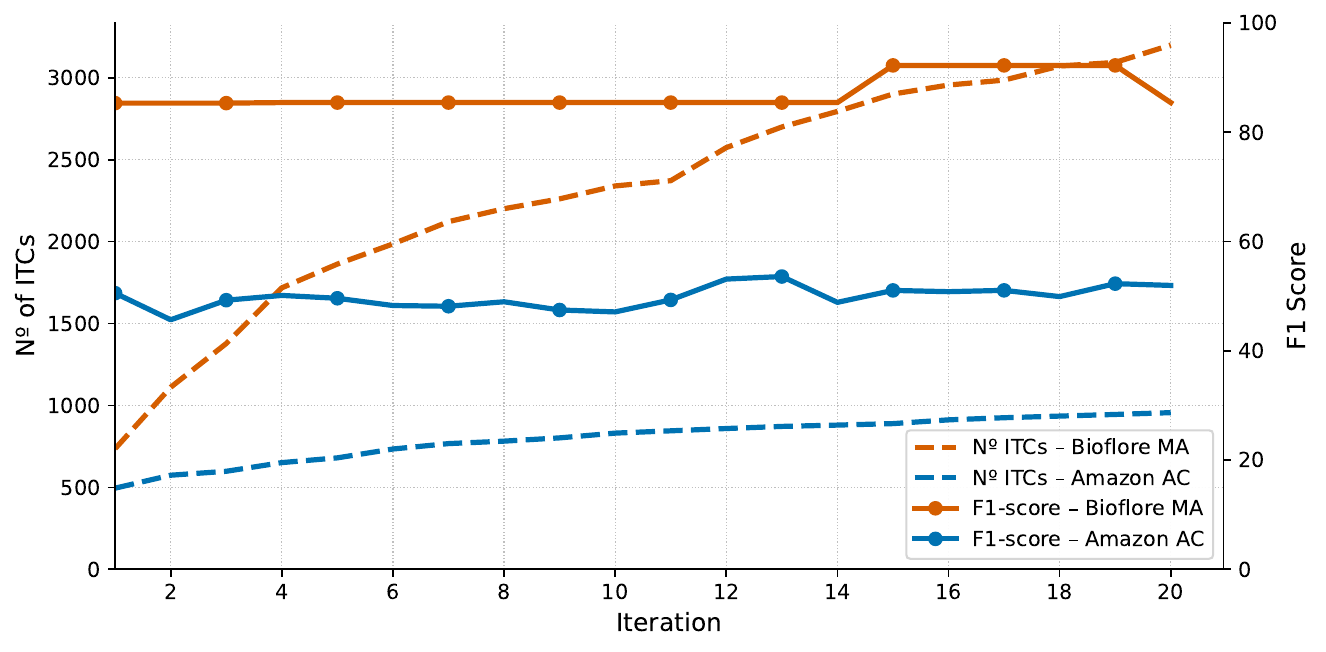}
    \caption{Number of individual tree crowns (ITCs) detected and macro-average F1-score across tree recruitment  iterations.}
    \label{fig:f1score_nr}
\end{figure*}

Figure \ref{fig:f1score_nr} presents the proposed method behaviour when considering how the number of Individual Tree Crowns (ITCs) detected and the F1-score in each datasets change, across 20 iterations. On the left y-axis, the Nº of ITCs is displayed, ranging from 0 to over 3000, while the F1-score is on the right y-axis, presented as a percentage from 0 to 100. The x-axis represents the Iteration count, from 0 to 20. Four distinct lines illustrate the key metrics for each dataset: a dashed orange line denotes Nº ITCs - Bioflore MA, a dashed blue line represents Nº ITCs - Amazon AC, a solid orange line with circles signifies F1-score - Bioflore MA, and a solid blue line with circles indicates F1-score - Amazon AC.

Analyzing the results depicted in Figure \ref{fig:f1score_nr}, a clear trend emerges for both datasets. For the Bioflore MA dataset (orange lines), the number of detected ITCs (orange dashed line) demonstrates a continuous increase across the tree recruitment  cycles, starting from approximately 700 ITCs at iteration 1 and escalating to over 3100 ITCs by iteration 20. Concurrently, the F1-score for Bioflore MA (solid orange line) exhibits stability and high performance, hovering around 85\% from iteration 1 to iteration 14, peaking at approximately 92\% between iterations 15 and 19, before a slight decrease to around 85\% by iteration 20. This indicates that the tree recruitment  approach expanded the number of identified ITCs without compromising the accuracy of the model, successfully maintaining a high F1-score.

 Similarly, for the Amazon AC dataset (blue lines), the number of detected ITCs (dashed blue line) also shows a steady growth, increasing from around 495 ITCs at iteration 1 to approximately 956 ITCs by iteration 20. The F1-score for Amazon AC (solid blue line), while generally lower than that of Bioflore MA, also experiences an overall upward trend, albeit with more pronounced fluctuations. It commences around 50\% and reaches its highest point near 54\% around iteration 13, stabilizing around 50\% towards the later iterations. For instance, between iterations 8 and 20, the F1-score remained steady, while the number of detected ITCs rose from 782 to 956, respectively. This suggests that even for a potentially more challenging or distinct dataset like Amazon AC, self-training contributes to an increased discovery of ITCs and an improvement in predictive accuracy over iterations, though the F1-score performance is not as high or stable as observed for Bioflore MA. The overall performance across both datasets confirms the effectiveness of tree recruitment  in iteratively improving the annotation process by identifying more high-confidence individual tree crowns while sustaining or improving model accuracy.

An observation, illustrated by the F1-score trajectories in Figure \ref{fig:f1score_nr}, is that an expectation of continuous, monotonic improvement in classification accuracy through successive tree recruitment  iterations is not always realistic. This outcome is a direct consequence of the self-training strategy employed, which, by its design, prioritizes the inclusion of only high-confidence samples into the inventory during each iteration. This selective inclusion means that the most challenging and ambiguous, i.e., borderline, samples where the model exhibits lower certainty, are systematically excluded from the model's direct training process. Because only confident candidates are recruited, these difficult cases, which are crucial for pushing the model's performance further, are left out of the automatic loop; in practice, they are the natural candidates to be delegated to human expert annotators. 

However, even relying on human annotation for these complex cases presents its own set of challenges. Within diverse environments, such as those represented by Bioflore MA and Amazon AC, the task of accurately discerning between species, often based on subtle morphological distinctions, can be difficult even for highly trained human experts. The current taxonomic complexities and the sheer volume of species further compound this challenge, making definitive and consistent labeling of borderline samples a challenging task. Consequently, as depicted by the F1-scores, the plateauing for Bioflore MA at a high level and the more fluctuating, modest gains for Amazon AC, the model's ability to improve beyond a certain threshold becomes constrained. This highlights a key limitation of high-confidence-based tree recruitment  when confronted with complex and ambiguous classification tasks, where the easy examples are quickly exhausted and the hard examples remain a persistent challenge for both machines and humans.

\begin{figure*}[t]
    \centering
    \includegraphics[width=1\textwidth]{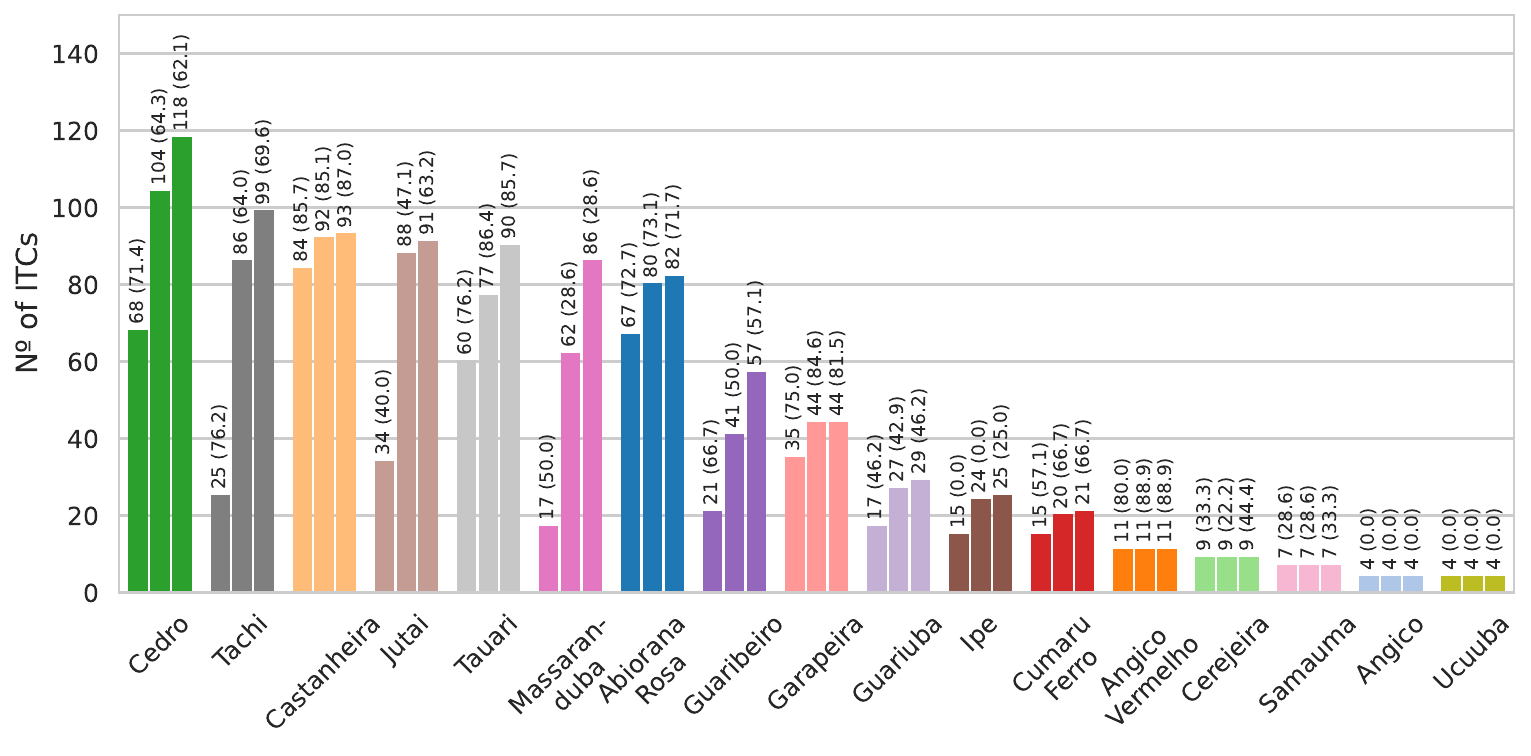}
    \caption{Number of predicted individual tree crowns (ITCs) in the 1st, 8th, and 13th iterations of the tree recruitment  cycle. A different bar color represents each species. The count of ITCs is displayed above each bar, with the corresponding F1-score in parentheses.}
    \label{fig:f1score_allspecies}
\end{figure*}

We conclude the  discussion with an analysis of the progression of ITCs added to the sample inventory during the self-training iterations. With this aim, Figure \ref{fig:f1score_allspecies} represents such a behavior and the average F1-score per species at iterations 1, 8, and 13 (shown in parentheses) on Amazon AC dataset. The primary objective of this figure is to illustrate the expansion of the training set driven by the recruitment criteria. Thus, rather than statistic representation of a dataset, the bar plot serves as a record of the model's acquisition priorities. It highlights how the proposed framework identified and incorporated ITCs into the inventory. This analysis focuses on the Amazon AC dataset, as it represents the most challenging scenario among the datasets evaluated in this study.

By inspecting the figure, can be observed that twelve species exhibited an increase in the number of ITCs from iteration 1 to 13, while five species (\textit{Angico Vermelho}, \textit{Cerejeira}, \textit{Samauma}, \textit{Angico}, and \textit{Ucuuba}) kept a constant count. \textit{Cedro} was the species with more ITCs (118) in the final iteration of the tree recruitment  process. On the other hand, the F1-score at the same time increased for the following species: \textit{Castanheira}, \textit{Jutai}, \textit{Tauari}, \textit{Garapeira}, \textit{Cumaru Ferro}, \textit{Angico Vermelho}, \textit{Cerejeira}, and \textit{Samauma} from the first to the 13th iteration. The largest improvement was observed in \textit{Jutai}, with an increase of 23.2 percentage points, followed by \textit{Cerejeira} with an increase of 11.1 percentage points. In contrast, six species, i.e., \textit{Ipê}, \textit{Massaranduba}, \textit{Cedro}, \textit{Guaribeiro}, \textit{Tachi}, and \textit{Abiorana Rosa}, experienced a decline in F1-score.

\FloatBarrier
\section{Conclusion}
\label{Conclusion}

This work presented a framework for reliable tree recruitment that progressively expands forest species inventories from a limited set of annotated individual tree crowns (ITCs) using high-resolution UAV imagery. The proposed framework combines multi-task semantic segmentation, confidence estimation, and morphology-aware validation to automatically recruit reliable tree crowns while preserving inventory quality throughout the expansion process.

Experimental results on two tropical forest datasets demonstrated that the progressive expansion of the inventory consistently improves tree species detection and classification compared with established semantic segmentation baselines, with marked gains for some species represented by few annotated crowns (e.g., \textit{Jutai}), although the rarest species, with fewer than ten annotated crowns, remained undetected by all methods. These results indicate that reliable tree recruitment provides an effective mechanism for reducing the dependence on extensive manual annotation while enabling scalable forest inventory expansion under scarce supervision.

Although the proposed framework performs well across distinct tropical forest environments, challenges remain in highly heterogeneous canopies, where overlapping crowns and complex forest structures can still affect segmentation quality and the reliability of automatically recruited samples. Future work will investigate the integration of multimodal remote sensing data, more advanced uncertainty estimation strategies, and the application of the proposed framework to larger and more diverse forest inventories.

Overall, this study demonstrates that progressive and reliable tree recruitment offers a practical pathway for automatically expanding forest species inventories, supporting biodiversity monitoring in data-scarce environments without requiring extensive new field campaigns or large-scale manual annotation.

\section*{Author Contributions}
L.F.S.L.: methodology, software, investigation, writing---original draft. P.J.S.V.: writing---original draft, writing---review and editing. M.P.F.: writing---review and editing. L.E.C.l.R., H.F.P.V., E.M.d.C.N., E.O.F., A.P.D.C., C.R.S., J.M.S.M.V., M.S.F., H.E.F.C.M.F., C.V.V., V.M.C., D.N.R., D.R.A.d.A. and P.G.M.: data curation, validation. D.A.B.O.: supervision. All authors have read and agreed to the submitted version of the manuscript.

\section*{Funding}

This study was funded by FAPERJ – Carlos Chagas Filho Foundation for the Support of Research in the State of Rio de Janeiro (Processes SEI 260003/013155/2024 and SEI 260003/006405/2024), by the National Council for Scientific and Technological Development (CNPq) (Brazil process number 420157/2023-9), by the Getulio Vargas Foundation Research and Applied Knowledge Network, by the Norway's Climate and Forest Initiative (NICFI, by World Wide Fund for Nature (WWF) Netherlands, by São Paulo Research Foundation (FAPESP) grant n. 2024/15211-1, and by the Open Domain Science project Forest Carbon Crime (Project Number: OCENW.M.21.203) of the Nederlandse Organisatie voor Wetenschappelijk Onderzoek (NWO).

\section*{Data Availability Statement}
The data presented in this study are not publicly available.

\section*{Conflicts of Interest}
J.M.S.M.V., M.S.F., H.E.F.C.M.F., C.V.V. and D.R.A.d.A. are affiliated with Bioflore. Bioflore did not fund this study and had no role in its design, in the analysis or interpretation of the data, in the writing of the manuscript, or in the decision to publish the results. The authors declare no conflicts of interest.

\bibliographystyle{IEEEtran}
\bibliography{refs}

\end{document}